\documentclass[sigconf]{acmart}

\usepackage{CJKutf8}
\usepackage[english]{babel}
\usepackage{booktabs}
\usepackage{graphicx}

\usepackage{tikz}
\usetikzlibrary{positioning, fit, mindmap, trees, calc,tikzmark,shapes}
\usetikzlibrary{shapes.arrows, fadings, automata,tikzmark,decorations.pathreplacing,patterns}
\usetikzlibrary{arrows.meta}
\usetikzlibrary{intersections}
\usepackage{tkz-euclide}
\usetikzlibrary{fit}

\usepackage{pgfplots}
\pgfplotsset{compat=1.14} 

\usepackage{amsmath}
\usepackage{amsfonts}
\usepackage{bm}

\usepackage{sistyle}
\SIthousandsep{,}

\usepackage{multirow}
\usepackage{subcaption}
\usepackage{adjustbox}
\usepackage[inline]{enumitem}

\usepackage{url}

\usepackage{xcolor}

\definecolor{tuatara}{RGB}{67, 67, 67}
\definecolor{aluminum}{RGB}{153,153,153}
\definecolor{silver}{RGB}{191,191,191}
\definecolor{platinum}{RGB}{228,228,228}
\definecolor{mercury}{RGB}{230,230,230}
\definecolor{gallery}{RGB}{240,240,240}
\definecolor{free_speech_aquamarine}{RGB}{0, 156, 114}
\definecolor{sun_shade}{RGB}{255, 144, 68}
\definecolor{fern}{RGB}{101,197,117}
\definecolor{french_blue}{RGB}{0, 112, 182}
\definecolor{matisse}{RGB}{25, 104, 167}
\definecolor{sushi}{RGB}{117, 168, 47}
\definecolor{shakespeare}{RGB}{85, 154, 193}
\definecolor{egg_shell}{RGB}{238, 234, 215}
\definecolor{carnation}{RGB}{245, 80, 86}
\definecolor{flamingo}{RGB}{237, 88, 85}
\definecolor{jet_stream}{RGB}{188, 214, 210}
\definecolor{jelly_bean}{RGB}{45, 126, 150}
\definecolor{tree_poppy}{RGB}{246, 154, 27}
\definecolor{deep_carmine_pink}{RGB}{236, 50, 67}
\definecolor{copper_rust}{RGB}{155, 64, 74}
\definecolor{midnight}{RGB}{0, 29, 50}
\definecolor{chilean_fire}{RGB}{215, 87, 44}
\definecolor{puerto_rico}{RGB}{94, 194, 166}
\definecolor{japanese_laurel}{RGB}{53, 116, 40}
\definecolor{fire_engine_red}{RGB}{206, 37, 51}
\definecolor{ku_crimson}{RGB}{243, 0, 25}
\definecolor{turmeric}{RGB}{211, 178, 76}
\definecolor{tahiti_gold}{RGB}{223, 102, 36}
\definecolor{outrageous_orange}{RGB}{255, 100, 45}
\definecolor{crusta}{RGB}{254, 127, 44}
\definecolor{safety_orange}{RGB}{254, 106, 0}
\definecolor{pigment_green}{RGB}{0, 175, 79}
\definecolor{jaffa}{RGB}{240, 131, 58}
\definecolor{jet_stream}{rgb}{0.69,0.61,0.85}
\definecolor{jelly_bean}{rgb}{0.47,0.32,0.66}
\definecolor{azalea}{RGB}{251, 196, 196}
\definecolor{sundown}{RGB}{249, 180, 181}
\definecolor{light_coral}{RGB}{244, 127, 123}
\definecolor{wewak}{RGB}{244, 143, 150}

\definecolor{biscay}{RGB}{44, 62, 80}
\definecolor{carmine_pink}{RGB}{231, 76, 60}
\definecolor{athens_gray}{RGB}{236, 240, 241}
\definecolor{celestial_blue}{RGB}{52, 152, 219}
\definecolor{curious_blue}{RGB}{41, 128, 185}
\definecolor{my_sin}{RGB}{255, 176, 59}
\definecolor{viridian}{RGB}{70, 137, 102}
\definecolor{tomato}{RGB}{255, 97, 56}
\definecolor{mountain_meadow}{RGB}{0, 163, 136}
\definecolor{padua}{RGB}{121, 189, 143}
\definecolor{killarney}{RGB}{56, 113, 66}
\definecolor{ocean_green}{RGB}{79, 176, 112}
\definecolor{pastel_green}{RGB}{107, 227, 135}
\definecolor{chinook}{RGB}{163, 232, 178}
\definecolor{cosmic_latte}{RGB}{222, 247, 229}
\definecolor{chateau_green}{RGB}{69, 191, 85}
\definecolor{RoyalBlue}{RGB}{69, 191, 85}

\renewcommand{\arraystretch}{1}

\makeatletter
\g@addto@macro\normalsize{%
  \abovedisplayskip 6pt plus 3pt minus 3pt%
  \belowdisplayskip \abovedisplayskip
  \abovedisplayshortskip 6pt plus3pt  minus3pt%
  \belowdisplayshortskip 6pt plus3pt minus3pt%
}

\makeatother

\copyrightyear{2018} 
\acmYear{2018} 
\setcopyright{acmcopyright}
\acmConference[CIKM '18]{2018 ACM Conference on Information and Knowledge Management}{October 22--26, 2018}{Torino, Italy}
\acmBooktitle{2018 ACM Conference on Information and Knowledge Management (CIKM'18), October 22--26, 2018, Torino, Italy}
\acmPrice{15.00}
\acmDOI{10.1145/XXXXXX.XXXXXX}
\acmISBN{978-1-4503-6014-2/18/10} 
\begin{document}

\title{Multi-Source Pointer Network for Product Title Summarization}
%\titlenote{Produces the permission block, and
%  copyright information}
%\subtitle{Extended Abstract}
%\subtitlenote{The full version of the author's guide is available as
%  \texttt{acmart.pdf} document}

\author{Fei Sun, Peng Jiang, Hanxiao Sun, Changhua Pei, Wenwu Ou, and Xiaobo Wang}
\affiliation{%
  \institution{Alibaba Group}
  \city{Beijing}
  \country{China}
}
\email{{ofey.sf,  jiangpeng.jp, hansel.shx, changhua.pch, santong.oww, yongshu.wxb}@alibaba-inc.com}

\begin{abstract}

In this paper, we study the product title summarization problem in E-commerce applications for display on mobile devices.
Comparing with conventional sentence summarization, product title summarization has some extra and essential constraints.
For example, factual detail errors or loss of the key information are intolerable for E-commerce applications.
Therefore, we abstract two more constraints for product title summarization:
\begin{enumerate*}[label=(\roman*)]
\item do not introduce irrelevant information;
\item retain the key information (\textit{e.g.}, brand name and commodity name).    
\end{enumerate*}
To address these issues, we propose a novel multi-source pointer network by adding a new \textit{knowledge encoder} for pointer network.
The first constraint is handled by \textit{pointer mechanism}, generating the short title by copying words from the source title.
For the second constraint, we restore the key information by copying words from the knowledge encoder with the help of the soft gating mechanism.
For evaluation, we build a large collection of real-world product titles along with human-written short titles. 
Experimental results demonstrate that our model significantly outperforms the other baselines.% and is very close to humans on ROUGE and METEOR scores.
Finally, online deployment of our proposed model has yielded a significant business impact, as measured by the click-through rate.
\end{abstract}

%todo: update
\begin{CCSXML}
<ccs2012>
 <concept>
<concept_id>10002951.10003317.10003347.10003357</concept_id>
<concept_desc>Information systems~Summarization</concept_desc>
<concept_significance>500</concept_significance>
</concept>
<concept>
<concept_id>10010178.10010179</concept_id>
<concept_desc>Artificial intelligence~Natural language processing</concept_desc>
<concept_significance>500</concept_significance>
</concept>
<concept>
<concept_id>10002951.10003317</concept_id>
<concept_desc>Information systems~Information retrieval</concept_desc>
<concept_significance>300</concept_significance>
</concept>
</ccs2012>
\end{CCSXML}

\ccsdesc[500]{Information systems~Summarization}
\ccsdesc[500]{Artificial intelligence~Natural language processing}
\ccsdesc[300]{Information systems~Information retrieval}

\keywords{Title Summarization; Pointer Network; Extractive Summarization}

\maketitle

\section{Introduction}
%todo: 给出commodity name的定义，case
%波司登羽绒服女2017时尚印花修身大毛领加厚中长款外套B		波司登大毛领羽绒服
%todo: 这样的数据 test：29596 45294，train：235256 362652

\begin{figure}[!htb]
  \centering  
  \begin{subfigure}[t]{.49\linewidth}
    \includegraphics[scale=0.097]{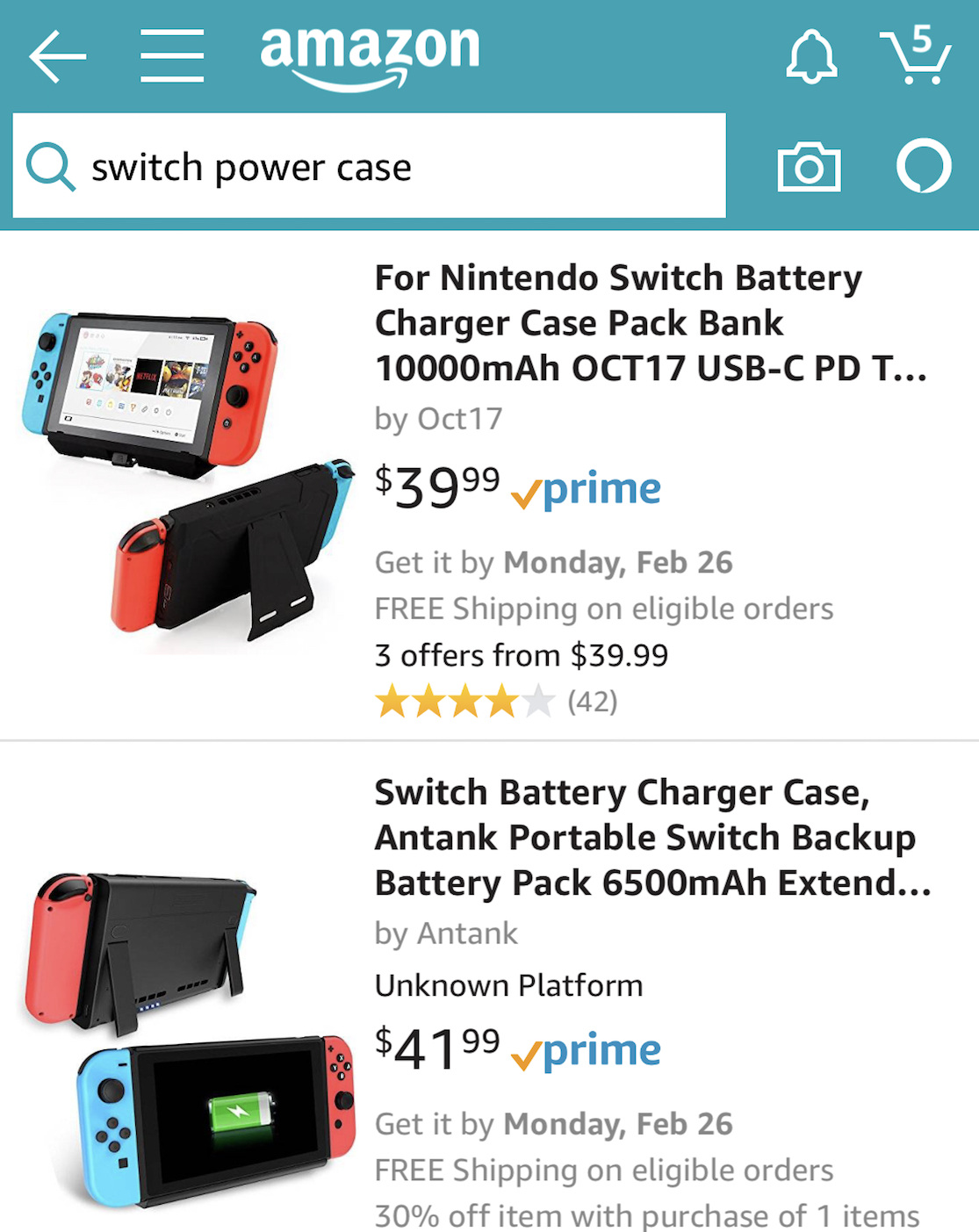}
    \caption{Search page on Amazon}
    \label{fig:amsearch}
  \end{subfigure}
   \begin{subfigure}[t]{.49\linewidth}
    \includegraphics[scale=0.097]{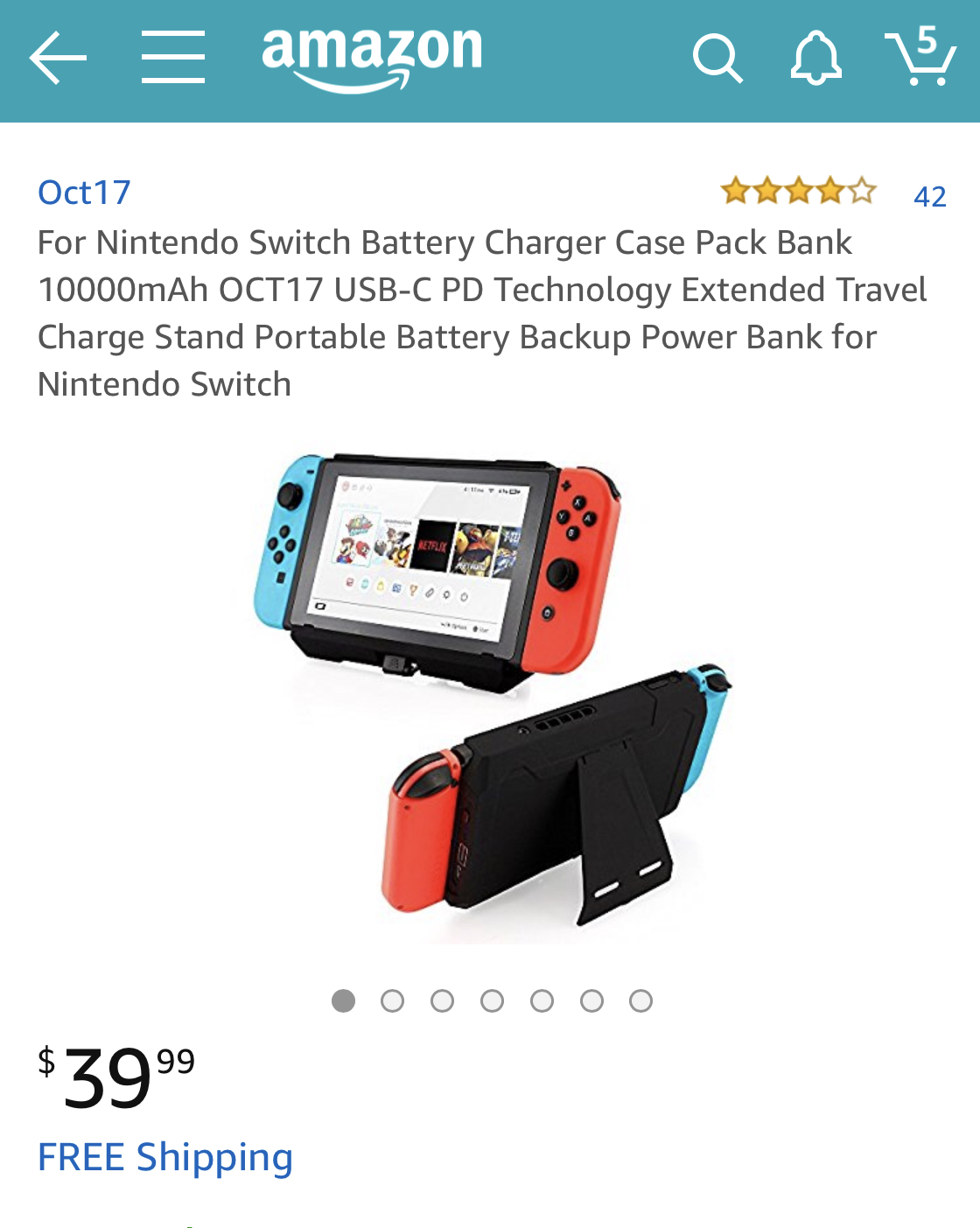}
    \caption{Detail page on Amazon}
    \label{fig:amdetail}
  \end{subfigure}
  \par\bigskip
  \begin{subfigure}[t]{.49\linewidth}
    \includegraphics[scale=0.087]{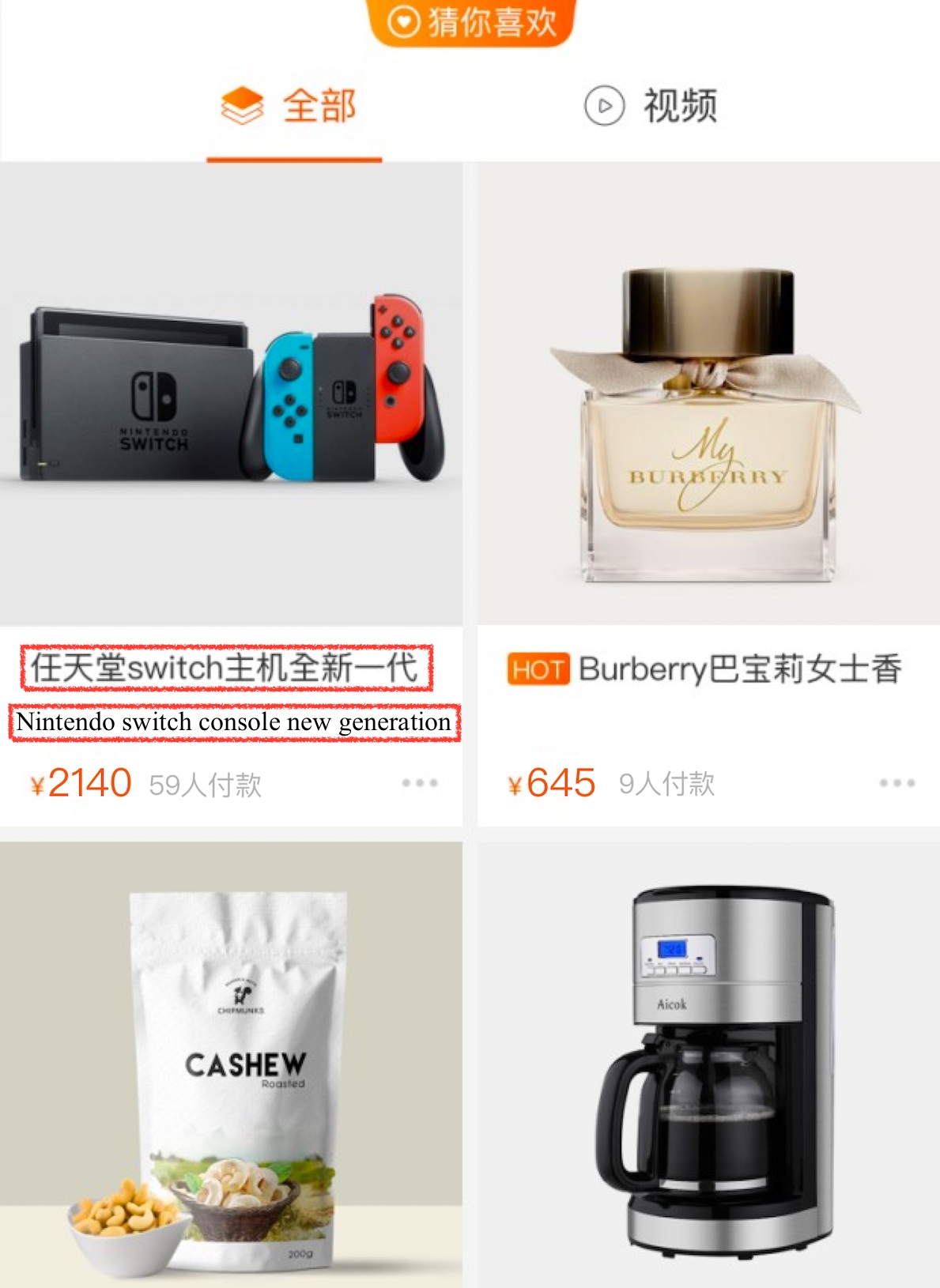}
    \caption{Recommendation page on Taobao}
    \label{fig:gul}
  \end{subfigure}
\begin{subfigure}[t]{.49\linewidth}
    \includegraphics[scale=0.087]{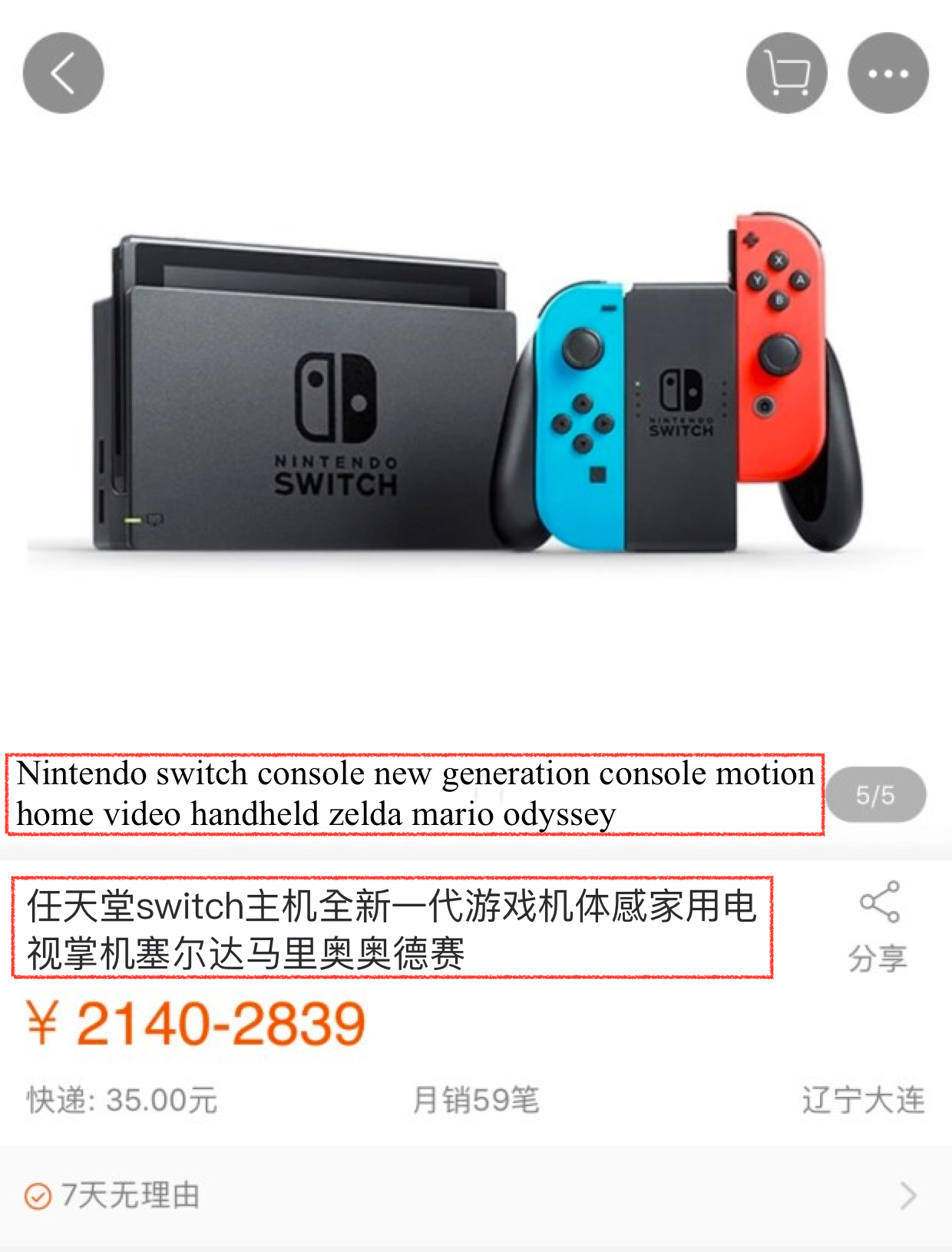}
    \caption{Detail page on Taobao}
    \label{fig:tbdetail}
  \end{subfigure}
  \caption{The product titles cannot display completely in corresponding pages on Amazon and Taobao iOS apps.
   %Sellers on E-commerce platforms tend to write redundant and lengthy titles for the sake of Search Engine Optimization. % (especially, customer to customer websites) 
   }
  \label{fig:example}
\end{figure}

%mobile phones play a ubiquitous role in our daily life.
Nowadays, more and more online transactions are made on mobile phones instead of on PCs. %\footnote{\url{http://www.alizila.com/11-11-gmv-shatters-record-new-retail-shows-promise}}.
However, some user interfaces in E-commerce applications are not optimized for mobile phones.
For screenshots in Figure~\ref{fig:amsearch} and~\ref{fig:gul}, the product titles cannot be fully displayed on two popular E-commerce apps. %  due to the limited screen size %even in multiple lines
In such case, user must go further into the detail page to see the full title of the product.
This really hurts users' browsing experience.
These redundant and lengthy product titles on the E-commerce platforms (especially, customer to customer(C2C) websites) are often produced by the sellers for the sake of Search Engine Optimization(SEO).
Although it is okay to display these lengthy titles on a PC's web browser, they are not suitable for displaying on the small screen of a mobile phone.
%it is not a good idea to use them on mobile phones.
Furthermore, products with a short and informative title may often better attract users’ attention and receive more clicks\footnote{\url{https://sellercentral.amazon.com/forums/message.jspa?messageID=2921001}} \cite{Wang:Multi:AAAI2018}.
Thus, generating a short and informative title for each product is an important and practical research problem in E-commerce.

We formalize this problem as product title summarization, a specific form of sentence summarization in the field of E-commerce.
However, comparing with the conventional sentence summarization \cite{Rush:Neural:EMNLP2015,Chopra:Abstractive:NAACL2016}, it has some extra and essential constraints.
For example, it is not a big problem to encounter incorrect factual details in the news summary system.
However, factual errors are intolerable for buyers and sellers on the E-commerce platforms.
Based on these considerations, we abstract two more stringent constraints for product title summarization:
\begin{enumerate*}[label=(\roman*)]
\item \textbf{Do not introduce irrelevant information};
\item \textbf{Retain the key information} (\textit{e.g.}, brand name and commodity name).	
\end{enumerate*}
%In fact, these two constraints are also implicitly reflected in sentence summarization.
These constraints are essential and explicit for product title summarization, since both the sellers and the consumers will be dissatisfied if we break any of them.

Firstly, sellers usually do not want the generated short titles mingled with the words not in their original titles.
This is because the words in original titles are usually carefully selected by the sellers and helpful to click-through rate \cite{Chakraborty:Stop:ASONAM2016}. 
Furthermore, it is totally unacceptable if the short title contains any
incorrect information, \textit{e.g.}, generating a wrong brand ``\textit{sony}'' in the short title for the product ``\textit{Nintendo Switch}''.
Although neural abstractive summarization methods have achieved great success in news or wiki-like articles, they still frequently generate incorrect factual details in the summaries~\cite{see:Get:ACL2017}.
This is why we do not employ neural abstractive methods in this work.

%That is to say, all the words in the compressed title are selected from the original title.
%Another consideration for choosing extractive way is that abstractive models may generate words that do not related to the original title.

Secondly, the generated short title should retain the key information in the original title.
%For example, losing the brand name ``\textit{Incase}'' or commodity name ``\textit{Sleeve}'' in the short title ``\textit{Incase ICON Sleeve for MacBook}'' will confuse the customers.
For example, it would be very confusing for customers if we lost the brand ``\textit{Incase}'' or commodity name ``\textit{Sleeve}'' in the short title ``\textit{Incase ICON Sleeve for MacBook}''.
Moreover, this is also unacceptable for the sellers on E-commerce platforms.

%Sequence-to-sequence (seq2seq) models ( Cho et al., 2014) have enjoyed great success in a variety of tasks such as machine translation, speech recognition, and text summarization.
% todo: pointer对于unknown
% todo: MS-Ptr的泛化能力，尤其对于线上没见过的数据

%Sequence-to-sequence (seq2seq) models ( Cho et al., 2014) have enjoyed great success in a variety of tasks such as machine translation, speech recognition, and text summarization.

In this paper, we propose a novel model named \textit{Multi-Source Pointer Network} (MS-Pointer) to explicitly model these two constraints. %extractively summarize the original redundant and lengthy product titles.
Specifically, for the first constraint, we model product title summarization as an extractive summarization problem using the Pointer Network \cite{Vinyals:Pointer:NIPS2015}, generating the short title by copying the words from the source title based on the attention mechanism \cite{Bahdanau:Neural:ICLR2015}.
That is to say, all the words in the compressed title are selected from the original title.

However, the pointer network cannot guarantee the generated short title containing the key information from the source title.
To tackle this issue, we extend the pointer network in a data driven way.
Specifically, we introduce a new \textit{knowledge encoder} for pointer network to encode the key information about the product. %can be seen as a memory, (\textit{e.g.}, brand name and commodity name) 
%At test time, the decoder can generate a word by copying it from not only the title encoder but also the knowledge encoder.
%Thus, MS-pointer can learn to decode the key information from the knowledge encoder in a data driven way.
At decoding time, MS-Pointer learns to copy different information from the corresponding encoders with the help of the soft gating mechanism.
In short, MS-Pointer can learn to decode the key information (\textit{e.g.}, brand name and commodity name) from the knowledge encoder in a data driven way.

For evaluation, we construct a large dataset containing \num{411267} product titles with corresponding human-written short titles from Taobao.com.
We compare our model with several abstractive and extractive baselines using both automatic and manual evaluations.
The results demonstrate that our model significantly outperforms several strong baselines.
In particular, on the brand retention test, MS-Pointer can correctly preserve more than 99\% brand names.
%In addition, further experiments on a smaller testset show that MS-Pointer performs very close to humans.
%By deploying the proposed model on an E-commerce mobile app, we witnessed improved online sales and better .
Finally, deployment of our MS-Pointer model on Taobao mobile app has yielded a significant business impact, as measured by the click-through rate.

% Finally, we deploy our DeepStyle search engine as a web-based application.
% e deployment of our solution has yielded a signi cant business impact, as measured by the conversion-rate.

\section{Related Work}
\label{sec:label}

In this section, we will briefly review the two lines of related works, \textit{i.e.}, sentence summarization and pointer mechanism.

\subsection{Sentence Summarization}
%text summarizaton is not the the focus of this paper
Generally, sentence summarization methods can be classified into two categories: abstractive methods and extractive methods.

%abstractive
Abstractive models generate the condensed sentence in a bottom-up way, \textit{i.e.}, creating a summary from scratch based on understanding the source text.
They build an internal semantic representation for the source text and then use natural language generation techniques to create a summary. %that is closer to what a human might express. 
The task of abstractive sentence summarization was formalized around the DUC-2003 and DUC-2004 competitions~\cite{Over:DC:PM2007}. 
Earlier studies mainly focused on~syntactic transduction \cite{Cohn:Sentence:Coling2008,Napoles:Paraphrastic:T2TW2011} and phrase-based statistical machine translation approach \cite{Banko:Headline:ACL2000,Wubben:Sentence:ACL2012,Cohn:Abstractive:TIST2013}.
Inspired by the success of neural machine translation \cite{Sutskever:Sequence:NIPS2014,Bahdanau:Neural:ICLR2015}, \citet{Rush:Neural:EMNLP2015} use convolutional models to encode the source, and a attentive feed-forward neural network to generate the summary.
Recently, \citet{Chopra:Abstractive:NAACL2016} extended \cite{Rush:Neural:EMNLP2015} with a attentive recurrent decoder.
Further, \citet{Nallapati:Abstractive:CoNLL2016A} proposed an RNN encoder-decoder architecture for summarization.

Comparing with abstractive methods, extractive models are more related to our work.
They assemble a summary by selecting a subset of important words from the original text.
Traditional approaches to this task focused on word deletion using rule-based \cite{Dorr:Hedge:DUC2003,Zajic:BBN:DUC2004} or statistical methods \cite{Knight:Statistics:AAAI2000,McDonald:Discriminative:EACL06,Filippova:Dependency:NLG2008,Galanis:extractive:NAACL2010,Woodsend:Title:EMNLP2010}.
More recently, deep learning models have also been applied to extractive sentence summarization.
\citet{Chen:Neural:ACL16} used RNN based encoder-decoder to learn a word extractor for extractive document summarization.
\citet{Filippova:Sentence:EMNLP2015} built a competitive sentence compression system via making the word deletion decision based on seqence-to-sequence framework. % with no linguistic features 
However, it is very difficult for the models based on deletion to deal with word reordering in the summarizations~\cite{Jing:2002:UHM:CL}.

\subsection{Pointer Mechanism} %(Ptr-Net)
%In particular, applying to machine translation task, (Luong et al., 2015) learns to point some words in source sentence and copy it to the target sentence, similarly to our method. However, it does not use atten- tion mechanism, and by having fixed sized soft max output over the relative pointing range (e.g., -7, ..., -1, 0, 1, ..., 7), their model (the Posi- tional All model) has a limitation in applying to more general problems such as summarization and question answering,
%  In question answering setting, (Hermann et al., 2015) have used placeholders on named entities in the context. However, the placeholder id is directly predicted in the softmax output rather than predict- ing its location in the context.

\textit{Pointer mechanism} was first introduced by \citet{Vinyals:Pointer:NIPS2015} to solve the problem of generating a sequence whose target dictionary varies depending on the input sequence. %whose tokens come from the input sequence
It uses attention mechanism as a pointer to select elements from the input sequence as output.
This allows it to generate previously unseen tokens.

Since being proposed, pointer mechanism has drawn more and more attention in text summarization \cite{Nallapati:Abstractive:CoNLL2016A,Miao:Language:EMNLP2016,see:Get:ACL2017,Paulus:Deep:ICLR2018}, machine translation \cite{Gulcehre:Pointing:ACL2016}, and dialogue generation \cite{Gu:Incorporating:ACL2016,Eric:Copy:EACL2017}, as it provides a potential solution for rare and out of vocabulary (OOV) words.
In addition, it has also been shown to be helpful for geometric problems \cite{Vinyals:Pointer:NIPS2015}, question answering \cite{Kadlec:Text:ACL2016,Wang:Machine:ICLR2017,Wang:Gated:ACL2017}, code generation \cite{Ling:Latent:ACL2016}, and language modeling \cite{Merity:Pointer:ICLR2017}.
It is also referred as \textit{copying mechanism} in text generation \cite{Gu:Incorporating:ACL2016,Eric:Copy:EACL2017,He:Generating:ACL2017}.
The key ideas of these works are very similar, extending pointer network in a soft \cite{Gu:Incorporating:ACL2016,see:Get:ACL2017} or hard \cite{Nallapati:Abstractive:CoNLL2016A,Gulcehre:Pointing:ACL2016} way to decide whether to generate a token from the predefined dictionary or from the input sequence.

%\subsection*{Others}
%todo: polish
\subsection*{Other related work}

Another related work is constrained sentence generation in dialogue systems \cite{Mou:Sequence:COLING2016,Xing:Topic:AAAI2017}.
\citet{Mou:Sequence:COLING2016} and \citet{Yao:Towards:EMNLP2017} only leveraged a single cue word in responses generation.
\citet{Xing:Topic:AAAI2017} used topic modeling to guide responses generation in conversation system.
However, none of these works meet the constraints of the task we studied in this paper.
%In contrast, our proposed model is more simple and general.
Perhaps, the closest work to ours is \cite{Wang:Multi:AAAI2018}.
\citet{Wang:Multi:AAAI2018} proposed a multi-task approach for product title compression using user search log data.
However, their work does not consider the second constraint we discussed in the introduction.

\section{Background}
In this section, we first review the sequence-to-sequence models which have been widely adopted for sentence summarization, and then introduce the pointer network for extractive summarization.

%todo: lstm and seq2seq are not the focus of the paper
\subsection{Sequence-to-Sequence(seq2seq) Model}
%todo: 补充
Recently, there has been a surge of work proposing to build the text summarization system within a seq2seq framework.
These models are usually composed of two RNNs, an encoder and a decoder.
The encoder maps the original text to a vector and the decoder transforms the vector to a summary.

Formally, denote the input text $\mathcal{S}=(w_1,w_2,\ldots,w_N)$  as a sequence of $N$ words, and the output sequence $\mathcal{Y}=(y_1,y_2,\ldots,y_M)$ as a $M$ words sequence. 
In the seq2seq framework, the source sequence $\mathcal{S}$ is converted into a fixed length vector $\bm{c}$ by the RNN encoder,
\begin{equation*}
	\begin{aligned}
		\bm{h}_t &= f(\bm{h}_{t-1}, \bm{w}_t) \\
		\bm{c} &= g\bigl(\{\bm{h}_1,\bm{h}_1,\ldots,\bm{h}_N\}\bigr) 
	\end{aligned}
\end{equation*}
where $\bm{w}_t$ is the word embedding of $w_t$, $\bm{h}_t$ is the RNN hidden state for word $w_t$ at step $t$, $f$ is the dynamics function of RNN unit, $\bm{c}$ is the so-called context vector, and $g$ is a function to summarize the hidden states $\bm{h}_t$ (\textit{e.g.}, a typical instance of $g$ is choosing the last state).
In practice, gated RNN alternatives such as LSTM~\cite{Hochreiter:LSTM:NC1997} or GRU~\cite{Cho:Learning:EMNLP2014} often perform much better than vanilla ones. 
Thus, in this work, we implement $f$ using LSTM~\cite{Hochreiter:LSTM:NC1997} which is parameterized as:
\begin{equation*}
	\begin{aligned}
		\bm{f}_t &= \sigma(\bm{W}_f \cdot [\bm{h}_{t-1}, \bm{x}_t] + \bm{b}_f) \\
		\bm{i}_t &= \sigma(\bm{W}_i \cdot [\bm{h}_{t-1}, \bm{x}_t] + \bm{b}_i) \\
		\bm{z}_t &= f_t \odot \bm{z}_{t-1} + \bm{i}_t \odot \tanh(\bm{W}_z \cdot [\bm{h}_{t-1}, \bm{x}_t] + \bm{b}_z) \\
		\bm{o}_t &= \sigma(\bm{W}_o \cdot [\bm{h}_{t-1}, \bm{x}_t] + \bm{b}_o) \\
		\bm{h}_t &= \bm{o}_t \odot \tanh (\bm{z}_t)
	\end{aligned}
\end{equation*}
where $\sigma$ is the sigmoid function, $\odot$ is element-wise multiplication, $\bm{i}, \bm{f}$, and $\bm{o}$ are respectively the \textit{input gate}, \textit{forget gate}, \textit{output gate}, $\bm{z}_t$ is the information stored in memory cells, all of which are the same size as the hidden vector $\bm{h}_t$, and all the non-linear operations are computed element-wise.
The subscripts for weight matrix and bias terms have the obvious meaning. 
For example $\bm{W}_f$ is the forget gate matrix, $\bm{b}_f$ is the bias term for the forget gate etc.

The decoder is to generate the target sequence based on the context vector $\bm{c}$ through the following dynamics process:
\begin{equation*}
	\begin{aligned}
		\bm{d}_{t} &= f(\bm{y}_{t-1}, \bm{d}_{t-1}, \bm{c})\\
		p(y_t = w|\mathcal{S}, y_{<t}) &= \phi(\bm{d}_{t})
	\end{aligned}
\end{equation*}
where $\bm{d}_{t}$ is the hidden state of the decoder at time step $t$, $y_t$ is the predicted target symbol at $t$ through function $\phi$, $\bm{y}_{t-1}$ is the word embedding of $y_{t-1}$, $y_{<t}$ denotes the history $(y_1,y_2,\ldots,y_{t-1})$.
In primitive decoder models, context vector $\bm{c}$ is the same for generating all the output words.
In practice, the attention mechanism~\cite{Bahdanau:Neural:ICLR2015} is usually adopted to dynamically change the context vector in order to pay attention to different parts of the input sequence at each step of the output generation.
%todo:加citation

\subsection{Pointer Network for Summarization}
\label{sec:pointer}

Unlike vanilla seq2seq models, pointer network \cite{Vinyals:Pointer:NIPS2015} uses the attention mechanism~\cite{Bahdanau:Neural:ICLR2015} as a pointer to select tokens from the input as output rather than picking tokens from a predefined vocabulary.
%Pointer mechanism is becoming an essential unit in seq2seq models like attention mechanism since it provide one potential solution for rare and OOV words (\textit{e.g.}, name entities).
This distinct characteristic makes pointer network very suitable for extractive  summarization.

Formally, given an input sequence $\mathcal{S}{=}(w_1,w_2,\ldots,w_N)$ of $N$ words, pointer network uses an LSTM as encoder to produce a sequences of \textit{encoder hidden states} $(\bm{h}_1,\bm{h}_2,\ldots,\bm{h}_N)$. 
At each step $t$, the decoder (a single-layer LSTM) produces the \textit{decoder hidden state} $\bm{d}_t$ using the word embedding of the previous word $y_{t{-}1}$ and the last step decode state $\bm{d}_{t{-}1}$. %\footnote{While training, this is the previous word of the reference summary; at test time it is the previous word emitted by the decoder.} 
Then, the \textit{attention distribution} $\bm{a}^{(t)}$ is calculated as in \cite{Bahdanau:Neural:ICLR2015}:
%then calculate the \textit{attention distribution} $\bm{a}_t$ as in \cite{Bahdanau:Neural:ICLR2015}: % :
\begin{equation*}
	\begin{aligned}
		u_{ti} &= \bm{v}^{\top} \tanh \left(\bm{W}_h \bm{h}_i + \bm{W}_d \bm{d}_t + \bm{b}_{\mathrm{attn}}\right) \\
		\bm{a}_t &= \mathrm{softmax} (\bm{u}_t)
	\end{aligned}
\end{equation*}
where softmax normalizes the vector $\bm{u}_t$ to be an distribution over the input position; $\bm{v}, \bm{W}_h, \bm{W}_d$, and bias term $\bm{b}_{\mathrm{attn}}$ are learnable parameters.

Considering a word $w$ may appear multiple times in the input sequence $\mathcal{S}$, we define the \textit{output distribution} of word $w$ by summing probability mass from all corresponding parts of the attention distribution, as in \cite{see:Get:ACL2017}:
\begin{equation*}
	p(y_t = w|\mathcal{S}, y_{<t}) = \sum_{i:w_i=w} a_{ti}
\end{equation*} %todo:解释a_ti
Finally, the training loss for step $t$ is defined as the negative log likelihood of the target word $w^{*}_t$ at that step:
\begin{equation*}
	\mathcal{L}_t = -\log p(y_t = w^{*}_t|\mathcal{S}, y_{<t})
\end{equation*}
%the overall loss for the whole sequence is the sum of loss $\mathcal{L}_t$ at each step $t$.

%todo: 增加数据的说明
\section{Multi-Source Pointer Network}
\label{sec:msp}

\setcounter{footnote}{1}
% free_speech_aquamarine
% carmine_pink
%padua
%celestial_blue
%shakespeare
%mountain_meadow
%matisse
\tikzset{
  emb/.style = {rectangle, fill=tomato, minimum width=1em, minimum height=2em},
  tit/.style = {text depth=.25ex, text height=1.5ex, inner sep=0, outer sep=0},
  barbox/.style={rectangle, inner sep=0, outer sep=0, fill=padua, anchor=south, minimum width=2mm},
  bemb/.style = {rectangle, fill=flamingo, minimum width=1em, minimum height=2em},
  bbarbox/.style={rectangle, inner sep=0, outer sep=0, fill=shakespeare, anchor=south, minimum width=2mm},
  dec/.style = {rectangle, fill=my_sin, minimum width=1em, minimum height=2em},
  tcv/.style = {rectangle, fill=carmine_pink, minimum width=1em, minimum height=2em},
  fd/.style={rectangle, inner sep=0, outer sep=0, fill=padua, anchor=south, minimum width=2mm},
}
%scale=0.6, every node/.style={transform shape}

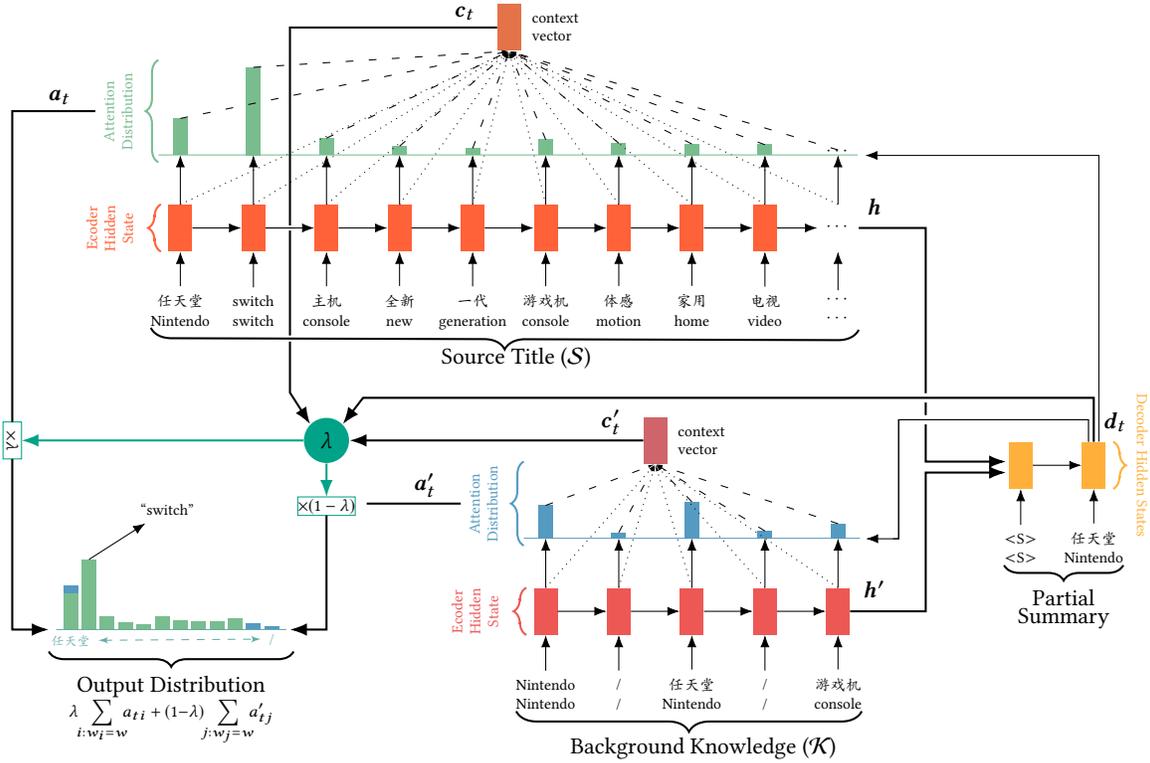
\begin{figure*}[htb]
  \centering
  \resizebox{0.87\textwidth}{!}{
  \begin{tikzpicture}[]
  \tikzstyle{every node}=[font=\scriptsize]

    \node[emb] (e1) at (0, 0) {};
    \foreach \x/\y in {1/2, 2/3, 3/4, 4/5, 5/6, 6/7, 7/8, 8/9} %, 9/10
    {
    \node[emb, right of=e\x, node distance=1cm] (e\y) {};
    }
    \node[emb, fill=white, right of=e9, node distance=1cm] (e10) {$\cdots$};
    \node[right of=e10, node distance=5mm, yshift=3mm] (ae10) {\normalsize$\bm{h}$};
    
    \node[below of=e1, node distance=1cm] (i1) {\begin{CJK*}{UTF8}{gkai}任天堂\end{CJK*}};
    \node[below of=e2, node distance=1cm] (i2) {switch};
    \node[below of=e3, node distance=1cm] (i3) {\begin{CJK*}{UTF8}{gkai}主机\end{CJK*}};
    \node[below of=e4, node distance=1cm] (i4) {\begin{CJK*}{UTF8}{gkai}全新\end{CJK*}};
    \node[below of=e5, node distance=1cm] (i5) {\begin{CJK*}{UTF8}{gkai}一代\end{CJK*}};
    \node[below of=e6, node distance=1cm] (i6) {\begin{CJK*}{UTF8}{gkai}游戏机\end{CJK*}};
    \node[below of=e7, node distance=1cm] (i7) {\begin{CJK*}{UTF8}{gkai}体感\end{CJK*}};
    \node[below of=e8, node distance=1cm] (i8) {\begin{CJK*}{UTF8}{gkai}家用\end{CJK*}};
    \node[below of=e9, node distance=1cm] (i9) {\begin{CJK*}{UTF8}{gkai}电视\end{CJK*}};
    %\node[below of=e10, node distance=1cm] (i10) {\begin{CJK*}{UTF8}{gkai}掌机\end{CJK*}};
    \node[below of=e10, node distance=1cm] (i10) {$\cdots$};
    \node[tit, below of=i1, node distance=0.25cm] (j1) {Nintendo};
    \node[tit, below of=i2, node distance=0.25cm] (j2) {switch};
    \node[tit, below of=i3, node distance=0.25cm] (j3) {console};
    \node[tit, below of=i4, node distance=0.25cm] (j4) {new};
    \node[tit, below of=i5, node distance=0.25cm] (j5) {generation};
    \node[tit, below of=i6, node distance=0.25cm] (j6) {console};
    \node[tit, below of=i7, node distance=0.25cm] (j7) {motion};
    \node[tit, below of=i8, node distance=0.25cm] (j8) {home};
    \node[tit, below of=i9, node distance=0.25cm] (j9) {video};
    %\node[tit, below of=i10, node distance=0.25cm] (j10) {handheld};
    \node[below of=i10, node distance=0.25cm] (j10) {$\cdots$};
    
    \foreach \x in {1,...,10}
    {
    \draw[-Latex] (i\x) edge (e\x) ;
    }
    
    \draw [thick,decorate,decoration={brace,amplitude=5pt}, tomato] ([xshift=-1mm] e1.south west) -- ([xshift=-1mm] e1.north west) node[midway,xshift=-0.7cm, rotate=90, align=center, tomato] (w) {Ecoder\\ Hidden\\State};
    
    \draw [thick,decorate,decoration={brace,amplitude=8pt,mirror}] (j1.south west) -- (j10.south east) node[midway,yshift=-0.4cm] (w) {\normalsize ~~~~Source Title ($\mathcal{S}$)};

    \node[barbox, minimum height=5mm] (a1) at ([shift={(90:1)}]e1) {};
    \node[barbox, minimum height=12mm] (a2) at ([shift={(90:1)}]e2) {};
    \node[barbox, minimum height=2.3mm] (a3) at ([shift={(90:1)}]e3) {};
    \node[barbox, minimum height=1.2mm] (a4) at ([shift={(90:1)}]e4) {};
    \node[barbox, minimum height=0.9mm] (a5) at ([shift={(90:1)}]e5) {};
    \node[barbox, minimum height=2.1mm] (a6) at ([shift={(90:1)}]e6) {};
    \node[barbox, minimum height=1.6mm] (a7) at ([shift={(90:1)}]e7) {};
    \node[barbox, minimum height=1.5mm] (a8) at ([shift={(90:1)}]e8) {};
    \node[barbox, minimum height=1.4mm] (a9) at ([shift={(90:1)}]e9) {};
    %\node[barbox, minimum height=2mm] (a10) at ([shift={(90:1)}]e10) {};
    \node[barbox, fill=white] (a10) at ([shift={(90:1)}]e10) {$\cdots$};
    
    \foreach \x in {1,...,10}
    {
    \draw[-Latex] (e\x) edge (a\x) ;
    }
    
    \node[emb, fill=padua!20!tomato, above of=a5, node distance = 1.7cm, xshift=0.5cm] (t_cv){}; %label={0:context vector}
    \node[right of=t_cv, node distance = 2em, align=left] (at_cv) {context\\ vector};
    \foreach \x in {1,...,10}
    {
      \draw[-{Latex[length=1mm, width=0.4mm]}, loosely dashed] (a\x.north) -> (t_cv.south) ;
      \draw[-{Latex[length=1mm, width=0.4mm]}, dotted] (e\x.north) -> (t_cv.south) ;
    }

    \draw [thick,decorate,decoration={brace,amplitude=5pt}, padua] ([xshift=-2mm, yshift=-1mm] a1.south west) -- ([xshift=-2mm, yshift=8mm] a1.north west) node[midway,xshift=-0.55cm, rotate=90, align=center, padua] (tad) {Attention\\Distribution};
    
    \path[-, draw=padua] ([xshift=-2mm] a1.south west) edge ([xshift=1mm] a10.south east);

    \node[bemb, below of=j6, node distance=4cm] (b1) {};
    \node[bemb, right of=b1, node distance=1cm] (b2) {};
    \node[bemb, right of=b2, node distance=1cm] (b3) {};
    \node[bemb, right of=b3, node distance=1cm] (b5) {};
    \node[bemb, right of=b5, node distance=1cm] (b4) {};
    \node[right of=b4, node distance=5mm, yshift=3mm] (ab4) {\normalsize$\bm{h}^{\prime}$};
    
    \node[below of=b1, node distance=1cm] (bi1) {Nintendo};
    \node[below of=b2, node distance=1cm] (bi2) {/};
    \node[below of=b3, node distance=1cm] (bi3) {\begin{CJK*}{UTF8}{gkai}任天堂\end{CJK*}};
    \node[below of=b5, node distance=1cm] (bi5) {/};
    \node[below of=b4, node distance=1cm] (bi4) {\begin{CJK*}{UTF8}{gkai}游戏机\end{CJK*}};
    \node[tit, below of=bi1, node distance=0.25cm] (jb1) {Nintendo};
    \node[tit, below of=bi2, node distance=0.25cm] (jb2) {/};
    \node[tit, below of=bi3, node distance=0.25cm] (jb3) {Nintendo};
    \node[tit, below of=bi5, node distance=0.25cm] (jb5) {/};
    \node[tit, below of=bi4, node distance=0.25cm] (jb4) {console};
    
    \node[bbarbox, minimum height=4.5mm] (ab1) at ([shift={(90:1)}]b1) {};
    \node[bbarbox, minimum height=0.8mm] (ab2) at ([shift={(90:1)}]b2) {};
    \node[bbarbox, minimum height=5mm] (ab3) at ([shift={(90:1)}]b3) {};
    \node[bbarbox, minimum height=1mm] (ab5) at ([shift={(90:1)}]b5) {};
    \node[bbarbox, minimum height=2mm] (ab4) at ([shift={(90:1)}]b4) {};
    
    \path[-, draw=shakespeare] ([xshift=-2mm] ab1.south west) edge ([xshift=2mm] ab4.south east);
    \draw [thick,decorate,decoration={brace,amplitude=8pt,mirror}] (jb1.south west) -- (jb4.south east) node[midway,yshift=-0.5cm] (w) {\normalsize ~~~~~Background Knowledge ($\mathcal{K}$)};
    \draw [thick,decorate,decoration={brace,amplitude=5pt}, flamingo] ([xshift=-1mm] b1.south west) -- ([xshift=-1mm] b1.north west) node[midway,xshift=-0.7cm, rotate=90, align=center, flamingo] (w) {Ecoder\\ Hidden\\State};
    \draw [thick,decorate,decoration={brace,amplitude=5pt}, shakespeare] ([xshift=-2mm, yshift=-1mm] ab1.south west) -- ([xshift=-2mm, yshift=6mm] ab1.north west) node[midway,xshift=-0.55cm, rotate=90, align=center, shakespeare] (bad) {Attention\\Distribution};
    
    \foreach \x in {1,...,5}
    {
    \draw[-Latex] (b\x) edge (ab\x) ;
    \draw[-Latex] (bi\x) edge (b\x) ;
    }
    \foreach \x/\y in {1/2, 2/3, 3/5, 5/4}
    {
    \draw[-Latex] (b\x) edge (b\y) ;
    }

    \node[dec, below of=j1, node distance=2cm, xshift=11.5cm] (d1) {};
    \node[dec, right of=d1, node distance=1cm] (d2) {};
    
    \node[below of=d1, node distance=1cm] (di1) {$<$S$>$};
    \node[below of=d2, node distance=1cm] (di2) {\begin{CJK*}{UTF8}{gkai}任天堂\end{CJK*}};
    \node[tit, below of=di1, node distance=0.25cm] (jd1) {$<$S$>$};
    \node[tit, below of=di2, node distance=0.25cm] (jd2) {Nintendo};
    
    \draw [thick,decorate,decoration={brace,amplitude=5pt, mirror}, my_sin] ([xshift=1mm] d2.south east) -- ([xshift=1mm] d2.north east) node[midway,xshift=0.4cm, rotate=-90, align=center, my_sin] (w) {Decoder Hidden States};
    
    \draw [thick,decorate,decoration={brace,amplitude=6pt,mirror}] (jd1.south west) -- (jd2.south east) node[midway,yshift=-0.6cm, align=center] (w) {\normalsize Partial\\ \normalsize Summary};
    
    \draw[-Latex] (d1) edge (d2) ;
    \foreach \x in {1,2}
    {
    \draw[-Latex] (di\x) edge (d\x) ;
    }
    \node[right of=d2, node distance=3mm, yshift=6mm] (ad2) {\normalsize$\bm{d}_t$};
    
    %\draw[-Latex, to path={-| (\tikztotarget)}] (e10) edge (d1) ;
    
    \node[inner sep=0, outer sep=0, left of=d1, node distance=2cm] (loc) {};
    
    \draw[-Latex, thick] (e10) -- ++(1.2, 0) -- ++(0, -3.2) -> ++(1.1, 0);
    \draw[-Latex, thick] (b4) -- ++(1.2, 0) -- ++(0, 1.9) -> ++(1.1, 0);
    
    \draw[-Latex] ($(d2.north)+(0.07,0)$) -- ++(0, 3.92) -- ++(-3.2, 0) ;
    
    \draw[-Latex, draw=white, double=black, double distance=\pgflinewidth, ultra thick] ($(d2.north)-(0.07,0)$) -- ++(0, 0.3) -- ++(-2.6, 0) -- ++(0, -1.63) -> ++(-0.5, 0);
    \draw[-Latex] ($(d2.north)-(0.07,0)$) -- ++(0, 0.3) -- ++(-2.6, 0) -- ++(0, -1.63) -> ++(-0.45, 0);
    
    \node[tcv, fill=flamingo!80!shakespeare,above of=ab2, node distance = 1.3cm, xshift=0.5cm] (b_cv){};
    \node[right of=b_cv, node distance = 2em, align=left] (ab_cv) {context\\ vector};
    \foreach \x in {1,...,5}
    {
      \draw[-{Latex[length=1mm, width=0.4mm]}, loosely dashed] (ab\x.north) -> (b_cv.south) ;
      \draw[-{Latex[length=1mm, width=0.4mm]}, dotted] (b\x.north) -> (b_cv.south) ;
    }
    
    \node[circle, fill=mountain_meadow, left of=b_cv, node distance=4.5cm, minimum size=0.618cm] (p) {\normalsize$\lambda$};
    
    \draw[-Latex, draw=white, double=black, double distance=\pgflinewidth, ultra thick] (d2.north) -- ++(0, 0.6) -- ++(-10, 0) -- (p.north east);
    \draw[-Latex, thick] (d2.north) -- ++(0, 0.6) -- ++(-10, 0) -> (p.north east); %my_sin
    \draw[-Latex, thick] (b_cv) -> (p);
    \draw[-Latex, thick] (t_cv) -- ++(-3, 0) -- ++(0, -5) -> (p.north west);
    
    \node[rectangle, minimum height=0.8em, minimum width=1.6em, inner sep=0, outer sep=0, draw=mountain_meadow, below of=p, node distance=0.9cm] (pb) {$\times (1-\lambda)$} ;
    \node[rectangle, minimum height=0.8em, minimum width=1.6em, inner sep=0, outer sep=0, draw=mountain_meadow, left of=p, node distance=4.3cm, rotate=-90] (pt) {$\times \lambda$} ;
    
    \node[barbox, below of=j1, node distance=4cm, minimum height=5mm, xshift=-15mm] (fd1) {}; %任天堂
    \node[bbarbox, minimum height=1.0mm] (fdb1) at ([shift={(90:2.5mm)}]fd1) {};
    \node[barbox, right=2.5mm of fd1.south, anchor=south, minimum height=9.6mm] (fd2) {}; %switch
    \node[barbox, right=2.5mm of fd2.south, anchor=south, minimum height=1.84mm] (fd3) {}; %主机
    \node[barbox, right=2.5mm of fd3.south, anchor=south, minimum height=0.96mm] (fd4) {};
    \node[barbox, right=2.5mm of fd4.south, anchor=south, minimum height=0.72mm] (fd5) {};
    \node[barbox, right=2.5mm of fd5.south, anchor=south, minimum height=1.76mm] (fd12) {}; %游戏机
    \node[barbox, right=2.5mm of fd12.south, anchor=south, minimum height=1.28mm] (fd6) {};
    \node[barbox, right=2.5mm of fd6.south, anchor=south, minimum height=1.2mm] (fd7) {};
    \node[barbox, right=2.5mm of fd7.south, anchor=south, minimum height=1.12mm] (fd8) {};
    \node[barbox, right=2.5mm of fd8.south, anchor=south, minimum height=1.6mm] (fd9) {};
    \node[bbarbox, right=2.5mm of fd9.south, anchor=south, minimum height=0.9mm] (fd10) {};
    \node[bbarbox, right=2.5mm of fd10.south, anchor=south, minimum height=0.4mm] (fd11) {};
    
    \path[-, draw=shakespeare!50!padua] ([xshift=-1mm] fd1.south west) edge ([xshift=1mm] fd11.south east);
    \node[below of=fd1, node distance=4mm, shakespeare!50!padua] (v1) {\begin{CJK*}{UTF8}{gkai}\tiny 任天堂\end{CJK*}} ;
    \node[below of=fd11, node distance=1.5mm, shakespeare!50!padua] (v11) {/} ;
    \draw[latex'-latex', dashed, shakespeare!50!padua] (v1) -- (v11);
    
    \draw [thick,decorate,decoration={brace,amplitude=6pt, mirror}] ([xshift=-2mm, yshift=-3mm] fd1.south west) -- ([xshift=2mm, yshift=-3mm] fd11.south east) node[midway,yshift=-8mm, align=center] (w) {\normalsize Output Distribution\\$\lambda\!\! \displaystyle\sum_{i:w_i{=}w}\!\!\! a_{ti} + (1{-}\lambda)\!\!\displaystyle\sum_{j:w_j{=}w}\!\!\! a^{\prime}_{tj}$};
    
    \node[above right=7mm of fd2] (v2l) {``switch''} ;
    \draw[-Latex] (fd2.north) -> (v2l);
    
    \draw[-Latex, mountain_meadow, thick] (p) -> (pb);
    \draw[-Latex, mountain_meadow, thick] (p) -> (pt);
    \draw[thick] (bad) -- ++(-1.6,0);
    \draw[thick] (tad) -- ++(-1.45, 0) -- (pt);
    \draw[-Latex, thick] (pb) -- ++(0, -1.69) -> ++(-0.5, 0);
    \draw[-Latex, thick] (pt) -- ++(0, -2.59) -> ++(0.5, 0);
    
    \node[left of=tad, node distance=8mm, yshift=2mm] (atad) {\normalsize$\bm{a}_t$};
    \node[left of=bad, node distance=8mm, yshift=2.5mm] (abad) {\normalsize$\bm{a}^{\prime}_t$};
    \node[left of=t_cv, node distance=6mm, yshift=2mm] (at_cv) {\normalsize$\bm{c}_t$};
    \node[left of=b_cv, node distance=6mm, yshift=2.5mm] (ab_cv) {\normalsize$\bm{c}^{\prime}_t$};
    
    \draw[-Latex, draw=white, double=black, double distance=\pgflinewidth, thick] (e2) -- (e3);
    \foreach \x/\y in {1/2, 2/3, 3/4, 4/5, 5/6, 6/7, 7/8, 8/9, 9/10}
    {
    \draw[-Latex] (e\x) edge (e\y) ;
    }
    
    \draw [thick,draw=white, double=black, double distance=\pgflinewidth, thick, decorate,decoration={brace,amplitude=8pt,mirror}] (j1.south west) -- (j10.south east);
    
  \end{tikzpicture}
  }
  \caption{Multi-source pointer network (MS-Pointer) with two encoders\protect\footnotemark. 
  The most distinctive characteristic of MS-Pointer is that it can \textit{copy} words from multiple encoders.
  At each decoding time step, a soft gating weight $\lambda \in [0,1]$ is calculated, which weights the probability of copying words from the source title, versus copying words from the background knowledge.
  The final output distribution (from which we make prediction) is weighted sum of attention distribution $\bm{a}_t$ and $\bm{a}^{\prime}_t$.}
  \label{fig:ms_ptr}
\end{figure*}

%\ref{fig:ms_ptr}
Although pointer network works very well in practice, it still loses the brand name or commodity name of product from time to time. 
We aim to endue the pointer network with the capacity retaining such key information in the generated short title.

To achieve this, in addition to the encoder for the source title, we introduce a new \textit{knowledge encoder}. 
It encodes the brand name and commodity name using an LSTM, just like what we have done with the source title.
At test time, the decoder can generate a short title by copying words from not only the title encoder but also the knowledge encoder.
In this way, the model can learn to decode the key information from the knowledge encoder in a data driven way.
The architecture of MS-Pointer model is shown in Figure~\ref{fig:ms_ptr}.

As Figure~\ref{fig:ms_ptr} shows, MS-Pointer combines the original title (``\textit{Nintendo switch console $\cdots$}'') and background knowledge (brand name ``\textit{Nintendo}'' and commodity name ``\textit{console}'') to produce the short title about the product ``\textit{Nintendo switch}''.\setcounter{footnote}{1}\stepcounter{footnote}\footnotetext{It is noteworthy that we use Chinese words here for convenience of presentation.
  In fact, our model is built on Chinese characters instead of Chinese words.}
Here, we simply concatenate the brand name and commodity name of the product as its background knowledge, using a separator ``/''\footnote{``/'' is also a separator between multi-language versions of the brane name, \textit{e.g.}, \textit{Nintendo}/\begin{CJK*}{UTF8}{gkai}任天堂\end{CJK*}.}.

Formally, for a product with source title $\mathcal{S}=(w_1,w_2,\ldots,w_N)$ and background knowledge $\mathcal{K}=(k_1,k_2,\ldots,k_M)$, we use LSTM to produce series of hidden states $(\bm{h}_1,\!\bm{h}_2,{\ldots},\bm{h}_N)$ and $(\bm{h}^{\prime}_1,\!\bm{h}^{\prime}_2,{\ldots},\bm{h}^{\prime}_M)$, respectively.
Next, we transform the final hidden states $\bm{h}_N$ and $\bm{h}^{\prime}_M$ into the initial state $\bm{d}_0$ of the decoder using rectified layer \cite{Glorot:Deep:AISTAT2011}: 
\begin{equation*}
	\bm{d}_0 = \mathrm{ReLU}\bigl(\bm{W}_f\cdot [\bm{h}_N, \bm{h}^{\prime}_M]\bigr)
\end{equation*}
where $\mathrm{ReLU} = \max(0, x)$, and $\bm{W}_f$ is learnable parameters. 
%todo: 整合两个的优点

For title encoder and knowledge encoder, we compute the attention distribution as follows: 
\begin{equation*}
	\begin{aligned}
	    u_{ti} &= \bm{v}^{\top} \tanh \bigl(\bm{W}_h \bm{h}_i + \bm{W}_d \bm{d}_t + \bm{b}_{\mathrm{attn}}\bigr) \\
		u_{tj}^{\prime} &= \bm{v}^{\prime\top} \tanh \bigl(\bm{W}^{\prime}_h \bm{h}^{\prime}_j + \bm{W}^{\prime}_d \bm{d}_t + \bm{b}^{\prime}_{\mathrm{attn}} \bigr) \\
		\bm{a}_t &= \mathrm{softmax} (\bm{u}_t), \quad \bm{a}_t^{\prime} = \mathrm{softmax} (\bm{u}_t^{\prime})
	\end{aligned}
\end{equation*}
where $\bm{a}_t$ is attention distribution for title encoder, $\bm{a}_t^{\prime}$ is attention distribution for knowledge encoder, $\bm{v}, \bm{v}^{\prime}, \bm{W}_h, \bm{W}^{\prime}_h, \bm{W}_d, \bm{W}^{\prime}_d, \bm{b}_{\mathrm{attn}}$, and $\bm{b}^{\prime}_{\mathrm{attn}}$ are parameters to be learned.
$\bm{d}_t$ is decoder hidden state at time step $t$, computed by:
\begin{equation*}
	\bm{d}_t = f(\bm{d}_{t{-}1}, \bm{y}_{t-1}, \bm{c}_{t-1}, \bm{c}^{\prime}_{t-1})
\end{equation*}%\setcounter{footnote}{2}
where $\bm{d}_{t-1}$ is decoder state at step $t{-}1$, $\bm{y}_{t{-}1}$ is the input of the decoder at step $t$ (the embedding of predicted target  word\footnote{During training, this is the embedding of the previous word in the reference summary. At test time, it is the embedding of the previous word emitted by the decoder.} $y_{t{-}1}$ at $t{-}1$), $f$ is a nonlinear function. 
Here, we use LSTM as $f$. 
$\bm{c}_{t-1}$ and $\bm{c}^{\prime}_{t-1}$ are context vectors for title encoder and knowledge encoder respectively, computed as:
\begin{equation*}
\bm{c}_t = \sum_{i} a_{ti} \bm{h}_i, \qquad \bm{c}^{\prime}_t = \sum_{i} a^{\prime}_{ti} \bm{h}^{\prime}_i
\end{equation*}
where, $a_{ti}$ is the weight of $\bm{a}_t$ at position $i$, and $a^{\prime}_{ti}$ is the weight of $\bm{a}^{\prime}_t$ at position $i$.

\subsection*{Output Distribution}
As shown in Figure~\ref{fig:ms_ptr}, in decoding, MS-Pointer tries to retain the key information with the help of the knowledge encoder.
Specifically, it learns to generate the brand name and the commodity name by picking words from the knowledge encoder.
To this end, we introduce a soft gating weight $\lambda$ to combine the attention distribution $\bm{a}_t$ and $\bm{a}^{\prime}_t$ as the final output distribution:
%we calculate the \textit{output distribution} of word $w$ based on not only the title's attention distribution $\bm{a}_t$ but also the knowledge's attention distribution $\bm{a}^{\prime}_t$:
\begin{equation*}
	p(y_t {=} w|\mathcal{S}, \mathcal{K}, y_{{<}t}) {=} \lambda\!\! \sum_{i:w_i=w}\!\!\! a_{ti} + (1{-}\lambda)\!\!\sum_{j:w_j=w}\!\!\! a^{\prime}_{tj}
\end{equation*}
%where $\lambda$ acts as a soft switch, choosing between copying a word from the source title or from the background knowledge.
Thus, the model can learn to copy words from different encoders by adjusting the gating weight $\lambda$.
Obviously, $\lambda$ should be able to automatically adjust according to the decode state $\bm{d}_t$, the decode input $\bm{y}_{t{-}1}$, the source title's context vector $\bm{c}_t$, and the background knowledge's context vector $\bm{c}^{\prime}_t$.
In this paper, we define it using sigmoid function:
%It is calculated from the decode state $\bm{d}_t$, the decode input $\bm{y}_{t{-}1}$, the source title's context vector $\bm{c}_t$, and the background knowledge's context vector $\bm{c}^{\prime}_t$:
\begin{equation*}
	\lambda = \sigma \bigl(\bm{w}_{d}^{\top}\bm{d}_t + \bm{w}_{y}^{\top}\bm{y}_{t{-}1} + \bm{w}_{c}^{\top}\bm{c}_t + \bm{w}_{c\prime}^{\top}\bm{c}^{\prime}_t \bigr)
\end{equation*}
where vector $\bm{w}_d, \bm{w}_{y}, \bm{w}_{c}$, and $\bm{w}_{c\prime}$ are parameters to be learned, and $\sigma(x) = 1/(1+\exp(-x))$. 

%It is clear to see that $\lambda$ can automatically adjust according to the decode state $\bm{d}_t$ and the input $\bm{y}_{t{-}1}$ at each step.
%We use the gating weight $\lambda$ to guide the decoder to copy words from different sources.
Here, the gating weight $\lambda$ works like a classifier to tell the decoder to extract different information from corresponding encoders. %\footnote{In practice, we use a trick to force the model to copy the brand from the knowledge encoder by masking the brand in the original title at the early stage of training.}.
At the first few steps, MS-Pointer usually produces a small $\lambda$.
In this way, our model can easily copy the brand name (\textit{e.g.}, \textit{Nintendo}) from the knowledge encoder.
After that, $\lambda$ will become larger to push the model to copy other \textit{modifier} information (\textit{e.g.}, \textit{motion} or \textit{video}) from the title encoder.
At last, $\lambda$ will become smaller again, so the knowledge encoder can help to decode the commodity information. 

Softmax is another feasible and general way to calculate the weights for each decoder.
In our test, there is no visible difference between sigmoid and softmax.
In this paper, we choose sigmoid as the case for easy explanation.
We also try to define the output distribution as the sum of all encoders' attention distribution, \textit{i.e.}, set $\lambda = 0.5$ constantly.
However, it works very poorly.
As~Figure~\ref{fig:ms_ptr} shown, the knowledge input is usually much shorter than the source title.
As a result, the decoder often generate the brand repeatedly, like ``\textit{NintendoNintendo$\ldots$}'', due to the higher probability of words in knowledge encoder.

Finally, we define the training loss as the the negative log likelihood of the target sequence:
\begin{equation*}
	\mathcal{L} = \frac{1}{T} \sum_{t=0}^{T} -\log p(y_t {=} w_t^*|\mathcal{S}, \mathcal{K}, y_{{<}t})
\end{equation*}
where $w_t^*$ is the target word at step $t$, $T$ is the length of the target sequence.

\section{Experiments}

%In this section, we first describe our experimental settings including the dataset, hyper-parameter selections, and baseline methods.

\subsection{Dataset Construction}
\label{sec:dataset}

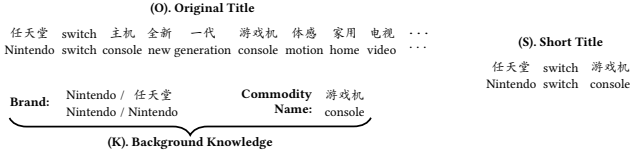
\begin{figure}
	\centering
	\resizebox{\linewidth}{!}{
	\begin{tikzpicture}[]
  \tikzstyle{every node}=[font=\scriptsize]
  
    \node[] (i1) {\begin{CJK*}{UTF8}{gkai}任天堂\end{CJK*}};
    \node[right of=i1, node distance=0.8cm] (i2) {switch};
    \node[right of=i2, node distance=0.7cm] (i3) {\begin{CJK*}{UTF8}{gkai}主机\end{CJK*}};
    \node[right of=i3, node distance=0.6cm] (i4) {\begin{CJK*}{UTF8}{gkai}全新\end{CJK*}};
    \node[right of=i4, node distance=0.7cm] (i5) {\begin{CJK*}{UTF8}{gkai}一代\end{CJK*}};
    \node[right of=i5, node distance=0.9cm] (i6) {\begin{CJK*}{UTF8}{gkai}游戏机\end{CJK*}};
    \node[right of=i6, node distance=0.75cm] (i7) {\begin{CJK*}{UTF8}{gkai}体感\end{CJK*}};
    \node[right of=i7, node distance=0.65cm] (i8) {\begin{CJK*}{UTF8}{gkai}家用\end{CJK*}};
    \node[right of=i8, node distance=0.6cm] (i9) {\begin{CJK*}{UTF8}{gkai}电视\end{CJK*}};
    \node[right of=i9, node distance=0.6cm] (i10) {$\ldots$};
    
    \node[tit, below of=i1, node distance=0.25cm] (j1) {Nintendo};
    \node[tit, below of=i2, node distance=0.25cm] (j2) {switch};
    \node[tit, below of=i3, node distance=0.25cm] (j3) {console};
    \node[tit, below of=i4, node distance=0.25cm] (j4) {new};
    \node[tit, below of=i5, node distance=0.25cm] (j5) {generation};
    \node[tit, below of=i6, node distance=0.25cm] (j6) {console};
    \node[tit, below of=i7, node distance=0.25cm] (j7) {motion};
    \node[tit, below of=i8, node distance=0.25cm] (j8) {home};
    \node[tit, below of=i9, node distance=0.25cm] (j9) {video};
    \node[below of=i10, node distance=0.25cm] (j10) {$\cdots$};
    
    \node[above of=i5, node distance=0.4cm] (s) {\textbf{(O). Original Title}};
    
    \node[below of=i1, node distance=1.1cm] (b) {\textbf{Brand:}};
    \node[below of=i2, node distance=1cm, xshift=0.2cm] (bi1) {Nintendo};
    \node[right of=bi1, node distance=0.5cm] (bi2) {/};
    \node[right of=bi2, node distance=0.5cm] (bi3) {\begin{CJK*}{UTF8}{gkai}任天堂\end{CJK*}};

    \node[below of=i8, node distance=1cm] (bi5) {\begin{CJK*}{UTF8}{gkai}游戏机\end{CJK*}};
    
    \node[right of=bi3, node distance=2cm, align=right, yshift=-0.1cm] (c) {\textbf{Commodity}\\\textbf{Name:}};

    \node[tit, below of=bi1, node distance=0.25cm] (jb1) {Nintendo};
    \node[tit, below of=bi2, node distance=0.25cm] (jb2) {/};
    \node[tit, below of=bi3, node distance=0.25cm] (jb3) {Nintendo};
    %\node[tit, below of=bi4, node distance=0.25cm] (jb5) {/};
    \node[tit, below of=bi5, node distance=0.25cm] (jb4) {console};
    
     \draw [thick,decorate,decoration={brace,amplitude=8pt,mirror}] ([xshift=1mm, yshift=-1.17cm]i1.south west) -- ([xshift=1mm]jb4.south east) node[midway,yshift=-0.4cm] (w) {\textbf{(K). Background Knowledge}};

    \node[right of=j10, node distance=1.5cm, yshift=-0.3cm] (d1) {\begin{CJK*}{UTF8}{gkai}任天堂\end{CJK*}};
    \node[right of=d1, node distance=0.8cm] (d2) {switch};
    \node[right of=d2, node distance=0.8cm] (d3) {\begin{CJK*}{UTF8}{gkai}游戏机\end{CJK*}};
    \node[tit, below of=d1, node distance=0.25cm] (de1) {Nintendo};
    \node[tit, below of=d2, node distance=0.25cm] (de2) {switch};
    \node[tit, below of=d3, node distance=0.25cm] (de3) {console};
    
    \node[above of=d2, node distance=0.4cm] (t) {\textbf{(S). Short Title}};

  \end{tikzpicture}}
  \caption{An example for the dataset.}
  \label{fig:dataset}
\end{figure}

For evaluation purposes, we build a new product title summarization dataset\footnote{\url{http://ofey.me/data/pts}} from Taobao.com, since there is no public benchmark dataset for our task yet.
Our proposed model requires two parts of data:
\begin{enumerate*}[label=(\roman*)]
\item original product titles and their corresponding short titles;
\item background knowledge about the products, \textit{i.e.}, brand name and commodity name.
\end{enumerate*}

In terms of $<$original title, short title$>$ pairs, we crawl the human-generated pairs from a product recommendation channel of the website.
The product titles and their corresponding short titles are manually written by professional editors (the corresponding short titles are rewritten in an extractive manner), thus suitable to be viewed as gold-standard for our task.
In terms of background knowledge, we collect these information from the corresponding fields in the database for each product. %online
Thus, the data set can be represented as ${<}O, K, S{>}$, where $O$ means products' original titles, $K$ means the background knowledge about the products, and $S$ represents the human-written short titles. 
A triplet example is presented in Figure~\ref{fig:dataset}.

We exclude the products whose short titles are longer than 10 Chinese characters since only 10 Chinese characters can be displayed in one line on mobile phones due to the screen size limitation.
%In addition, the products whose original titles miss the brand or short titles are not extractive (\textit{i.e.}, containing the words not appeared in the original title) are also filtered out.
Eventually, we get a dataset with \num{411267} pairs in 94 categories.
Table~\ref{tab:dataset} provides the detailed statistics about this dataset.
Finally, we randomly stratified split the dataset into a training set (80\%, \num{329248} pairs), a validation set (10\%, \num{41031} pairs), and a test set (10\%, \num{40988} pairs) by preserving the percentage of samples for each category. 

\begin{table}
  \centering
  \caption{The statistics of the data set. All lengths are counted by Chinese characters, and English word is counted as Chinese character.} 
  \label{tab:dataset}
  \begin{adjustbox}{max width=\linewidth}
  \begin{tabular}{l r} \toprule
    Dataset size & \num{411267} \\
    Number of category & \num{94}\\
    Avg. length of original titles &  25.42\\
    Avg. length of short titles & 7.77 \\
    Avg. length of background knowledge & 5.91 \\
    \bottomrule
  \end{tabular}
  \end{adjustbox}
\end{table}

\subsection{Baselines}

To verify the effectiveness of our proposed model, we compare it with two classes of baselines.

The abstractive methods including:
\begin{itemize}
\item Vanilla sequence-to-sequence (\textbf{Seq2Seq-Gen}) is a basic en-coder-decoder model based on LSTM unit \cite{Hochreiter:LSTM:NC1997} and attention mechanism \cite{Bahdanau:Neural:ICLR2015}.
\item Pointer-Generator (\textbf{Ptr-Gen}) \cite{see:Get:ACL2017} is a hybrid model combing Seq2Seq-Gen with pointer network. 
Besides copying words from the input, Ptr-Gen can also generate words from the predefined vocabulary.
\end{itemize}

The extractive methods including:
\begin{itemize}
\item Truncation (\textbf{Trunc.}) is the simplest baseline for product title summarization, where the words are kept in their original order until the limit is reached.
It is the practical solution in most E-commerce applications (\textit{e.g.}, Amazon, eBay, and Taobao).
%In most E-commerce applications (\textit{e.g.}, Amazon.com, eBay, and Taobao), product titles are simply truncated to adapt to the limitations.
\item \textbf{TextRank} \cite{Mihalcea:2004:EMNLP} is a keyword extraction framework.
It builds an automatic summary by extracting keywords or sentences from the text according to their scores computed by an algorithm similar to PageRank.
\item \textbf{Seq2Seq-Del} is introduced by Filippova et al.~\cite{Filippova:Sentence:EMNLP2015} to compress the sentence by deletion in a seq2seq framewaork.
Different from Seq2Seq-Gen generating words for summarization at each step, Seq2Seq-Del predicts the binary label (\textit{i.e.}, delete or retain) for each words in the original title at each decode steps\footnote{The decoder's input for Seq2Seq-Del is also the original title. This is different from the vanilla sequence-to-sequence models.}.
\item \textbf{LSTM-Del} is a standard sequence labeling system. 
It can be seen as a simplified version of Seq2Seq-Del since LSTM-Del performs the binary labeling based on the encoder's outputs. %todo: 解释清楚
\item Pointer network (\textbf{Ptr-Net}) \cite{Vinyals:Pointer:NIPS2015} is an extractive summarization baseline as we introduced in Section~\ref{sec:pointer}.
\item \textbf{Ptr-Concat} is the pointer network with concatenated input (\textit{i.e.}, concatenating the background knowledge with the source  title). We implement this baseline to better evaluate the effectiveness of our proposed model.
\end{itemize}

For model $x$ built on RNN architecture, we implement two versions based on  unidirectional and bidirectional LSTM unit, indicated as $x_{\text{uni}}$ and $x_{\text{bi}}$, respectively. %polish
Generally, we omit the subscript to refer to the model, not a specific implementation.
%However, in some case without ambiguity in the context, the model name without subscript also stands for its bidirectional version.

\subsection{Experiment Settings}

We implement all the models in \texttt{Tensorflow}\footnote{\url{https://www.tensorflow.org}} except Trunc. and TextRank.
For TextRank, we adopt the implementation in an open-source Python library \texttt{SnowNLP}\footnote{\url{https://github.com/isnowfy/snownlp}}.
For Ptr-Gen\footnote{\url{https://github.com/abisee/pointer-generator}}, we modify the IO of the code released by the authors to fit our dataset.
For all RNN encoder-decoder models, we use 128-dimensional word embeddings and 256-dimensional hidden states for LSTM units in both encoder and decoder.
For bidirectional implementations, we linearly transform the forward hidden states and backward hidden states into 256-dimensional states.

For all experiments, the word embeddings are initialized using normal distribution $\mathcal{N}(0, 10^{-8})$ and learned from scratch.
Other learnable parameters are initialized in the range $[-0.02, 0.02]$ uniformly.
We train all the models using Adagrad \cite{Duchi:Adaptive:JMLR2011} with learning rate $0.15$ and an initial accumulator value of $0.1$.
The gradient is clipped \cite{Pascanu:difficulty:ICML2013} when its $\ell_2$ norm exceeds a threshold of $2$.
We do not use any form of regularization in all experiments.
All the models are trained on a single Tesla M40 GPU with a batch size of 128.
The validation set is used to implement early stopping and tune the hyperparameters.
At test time, we decode the short titles using beam search with beam size 4 and maximum decoding step size\footnote{The lsat token is invisible token \textit{$<$EOS$>$}. So the real max length of generated short title is 10, as in training setting.} 11.
For Seq2Seq-Del and LSTM-Del, we keep all the words with positive predicted label (\textit{e.g.}, retain).

In this work, we try to minimize the preprocessing on the dataset.
All models are implemented based on Chinese characters for the following considerations:
\begin{enumerate*}[label=(\roman*)]
\item avoid the effect of Chinese word segmentation error;
\item easily control the output length;
\item models based on Chinese characters perform better.
\end{enumerate*}
We only ignore the numbers that do not appear in the products' brand name.
The product model (\textit{e.g.}, \textit{PS4}, \textit{DDR4}, or \textit{XXL}), the numbers in brand (\textit{e.g.}, \textit{7.Up} or \textit{5.11}) or punctuations in brand (\textit{e.g.}, \textit{Coca-Cola}, \textit{J.crew}, or \textit{Kiehl's}) are all kept during training and testing. %\textit{7} in \textit{7 For All Mankind}
During training, we keep the tokens occurring greater than 2 times, resulting
in a vocabulary of \num{47110} tokens (\num{4260} Chinese characters and \num{42850} other tokens).
For models based on pointer mechanism, we add the term \textit{$<$EOS$>$} to the end of the source title, so they can terminate at the decoding time.

\subsection{Automatic Evaluation}
\begin{table*}
  \centering
  \caption{BLEU, ROUGE (F$_1$), and METEOR scores on the test set. Baselines on the top group are abstractive, while those in the following two groups are extractive. 
  %All ROUGE scores have a 95\% confidence interval of at most $\pm0.25$. 
  Bold scores are the best overall.}
  \label{tab:full}
  \begin{adjustbox}{max width=\textwidth}
  %\resizebox{0.5\textwidth}{!}{
  \begin{tabular}{l c c c c c c c } \toprule
    &  BLEU-1 & BLEU-2 & BLEU-4 & ROUGE-1 & ROUGE-2 & ROUGE-L & METEOR \\ \midrule
    Seq2Seq-Gen$_{\text{uni}}$ & 69.26 & 60.51 & 43.79 & 68.69 & 50.57 & 68.38 & 37.04 \\
    Seq2Seq-Gen$_{\text{bi}}$ & 69.87 & 61.34  & 44.71 & 68.87 & 51.09 & 68.56 & 37.43 \\
    Ptr-Gen$_{\text{uni}}$ & 72.70 & 64.54 & 48.87 & 71.91 & 55.37 & 71.63 & 38.48 \\
    Ptr-Gen$_{\text{bi}}$ & 73.13  & 65.15 & 50.04 & 72.80 & 56.32 & 72.55 & 38.86 \\ \midrule
    Trunc. & 44.31 & 36.26 & 23.58 & 45.94 & 30.16 & 44.35 & 26.57 \\
    TextRank & 38.74 & 30.50 & 17.17 & 39.59 & 25.42 & 34.10 & 18.11 \\
    LSTM-Del$_{\text{uni}}$ & 63.67 & 52.28 & 33.52 & 65.23 & 43.61 & 64.35 & 33.87 \\
    LSTM-Del$_{\text{bi}}$ & 67.10 & 56.32 & 37.63 & 67.34 & 46.93 & 66.59 & 34.77 \\
    Seq2Seq-Del$_{\text{uni}}$ & 66.15  & 56.21 & 38.47 & 68.49  & 48.53 & 67.70  & 35.84 \\
    Seq2Seq-Del$_{\text{bi}}$  & 67.46 & 57.01  & 38.65 & 68.88 & 49.64 & 67.99 & 36.95  \\
    Ptr-Net$_{\text{uni}}$ & 73.88 & 65.81 & 50.73 & 73.40 & 56.93 & 73.13 & 39.10 \\
    Ptr-Net$_{\text{bi}}$ & 74.19  &  66.30 & 51.12 & 74.25 & 57.95 & 74.11 & 39.35 \\  \midrule
    Ptr-Concat$_{\text{uni}}$ & 74.25  & 66.09 & 50.86  & 74.07  & 58.13 & 73.89 & 39.25\\
    Ptr-Concat$_{\text{bi}}$ & 74.53  & 66.51 & 51.43 & 74.49 & 58.67 & 74.26 & 39.41 \\
    MS-Pointer$_{\text{uni}}$  & 75.11 & 67.17 & 52.55  & 75.15 & 59.62 & 74.96 & 39.91 \\
    MS-Pointer$_{\text{bi}}$  & \textbf{75.57}  & \textbf{67.72}  & \textbf{53.06}  & \textbf{75.69} & \textbf{60.29} & \textbf{75.45} & \textbf{40.25}\\
    \bottomrule
  \end{tabular}%}
  \end{adjustbox}
\end{table*}

To automaticlly evaluate the performance of different models, we leverage three standard metrics: BLEU \cite{Papineni:Bleu:ACL2002}, ROUGE \cite{Lin:ROUGE:2004}, and METEOR \cite{Banerjee:METEOR:MTSumm2005}.
The BLEU metric is originally designed for machine translation by analyzing the co-occurrences of $n$-grams between the candidate and the references.
For BLEU metric, we report BLEU-1, BLEU-2, and BLEU-4 here.
The ROUGE metric measures the summary quality by counting the overlapping units (\textit{e.g.}, $n$-grams) between the generated summary and reference summaries.
Following the common practice, we report the F$_1$ scores for ROUGE-1, ROUGE-2, and ROUGE-L. %(``L'' stands for \textit{longest common subsequence}).
The METEOR metric was introduced to address several weaknesses in BLEU.
It is calculated based on an explicit alignment between the unigrams in the candidate and references.
In this work, the METEOR scores are evaluated in exact match mode (rewarding only exact matches between words).
We obtain the BLEU and METEOR scores using the \texttt{nlg-eval}\footnote{\url{https://github.com/Maluuba/nlg-eval}} packages, and ROUGE scores using \texttt{pythonrouge}\footnote{\mbox{\url{https://github.com/tagucci/pythonrouge}}} package.

% \cite{Shikhar:Relevance:arxiv2017} 

%todo: attention的position问题

Table~\ref{tab:full} presents the results on seven automatic metrics.
As expected, Trunc.~and TextRank perform the worst on all metrics.
For Trunc., the reason is it has no ability to extract the informative words (\textit{e.g.}, commodity name like ``\textit{console}'' or ``\textit{handheld}'') from the tail of the product title.
For TextRank, this is because the algorithm cannot extract meaningful keyword in such short text.
Deletion based summarization models (\textit{i.e.}, Seq2Seq-Del and LSTM-Del) perform significantly better than TextRank and Trunc.~with a large gap.
However, they are much worse than other seq2seq models like Ptr-Net and MS-Pointer.
This is mainly due to the reordering phenomenon in short titles.
Take the first case in Table~\ref{tab:cs} as an example, the short title changes the order between ``\textit{MOD-X}'' and ``\textit{Nintendo Switch}''.
In fact, we find that more than half of the data (65.39\% in training set and  65.81\% in test set) adjust the word order to produce more fluent short titles.
Seq2Seq-Del outperforms LSTM-Del mainly because it can leverage more information from the entire title representation under encoder-decoder framework. 

Comparing different groups, it is easy to see that extractive models based on pointer mechanism perform better than abstractive models, especially significantly better than Seq2Seq-Gen.
This demonstrates that pointer mechanism is very suitable for extractive summarization.
An interesting phenomenon is that Ptr-Gen performs very close to Ptr-Net.
This is because, as a hybrid model combing Seq2Seq-Gen with Ptr-Net, Ptr-Gen can degenerated into Ptr-Net on an extractive dataset.
However, Ptr-Gen still make some factual mistakes occasionally, such as generating wrong brand name, replacing an uncommon (but in-vocabulary) word with a more-common alternative.

%ptr-concat与point 差异不大
%As we can see that models built on pointer mechanism are very strong baselines.
The results of the last group indicate that the background knowledge is helpful for product title summarization. 
Moreover, the gaps between MS-Pointer and Ptr-Concat demonstrate that our proposed model is more flexible and effective in modeling such knowledge. %todo: 更有力的解释
Finally, the improvements from these baselines to MS-Pointer are statistically significant using a two-tailed t-test ($p {<} 0.05$).
%Finally, MS-Pointer$_{\text{bi}}$ beats all other baselines on all metric as the bold scores shown in table~\ref{tab:full}.

As shown in the table~\ref{tab:full}, comparing with unidirectional LSTM, the bidirectional LSTM only bring a very limited improvement for each model.
This is largely because our input is very short and the decoder can only use unidirectional model.
Considering the superiority of the bidirectional models, we only report the results of bidirectional models (omitting the subscript for convenience) in the following evaluations.

\subsection{Brand Retention Test}
\label{sec:brand}

Besides the standard metric for text generation, we also test whether the model retains the brand in the source title.
Compared with normal sentence summarization task, retaining brand name is a particular and essential requirement for product title summarization.
The evaluation metric for this task is the error rate of the brand names in the generated short titles. %todo: polish
Besides the \textit{offline} testset used before, we also build a new \textit{online} testset by randomly sampling \num{140166} product titles with brand names from Taobao.com.

\begin{table}
  \centering
  \caption{Results of brand retention experiment. Bold scores are the best}
  \label{tab:brand}
 % \begin{adjustbox}{max width=0.5\textwidth}
  %\resizebox{0.5\textwidth}{!}{
  \begin{tabular}{l c c } \toprule
    &  offline & online \\ \midrule
   Seq2Seq-Gen & 9.82\% & 28.55\% \\
   Ptr-Gen & 2.85\% & 15.03\% \\ \midrule
   Trunc. & 2.42\% & 6.31\% \\
   TextRank & 92.43\%  & 93.08\%  \\
   LSTM-Del & 5.97\% & 22.31\% \\
   Seq2Seq-Del & 4.71\% & 19.76\% \\
   Ptr-Net &  2.54\% &  13.82\% \\
   Ptr-Concat & 1.33\% & 6.48\% \\
   MS-Pointer & \textbf{0.13\%} &  \textbf{2.89\%} \\
    \bottomrule
  \end{tabular}%}
  %\end{adjustbox}
\end{table}

Table~\ref{tab:brand} shows the results of each model on two testsets.
As expected, TextRank achieves the worst results in this task because of its ignorance of semantics.
Trunc.~performes very well on this task because the brand name usually appears at the head of the title.
Moreover, our MS-Pointer significantly outperforms the other baselines, especially on \textit{online} testset.
This demonstrates that introducing the knowledge encoder can help the pointer network to better retain the brand name. %in short title.
In addition, the results on the \textit{online} testset also indicates that our MS-Pointer has a stronger generalization ability.
The additional bad case analysis indicates that MS-Pointer's 2.89\% error rate on \textit{online} testset is largely due to the unknown (OOV) words.
It is not easy for pointer mechanism to copy a right word when the input contains several OOV words, since all these OOV words share the same embedding for the symbol $<$\textit{UNK}$>$.
In practice, we reduce the error rate of MS-Pointer from 2.89\% to \textbf{0.56}\% on \textit{online} testset by mapping each OOV word to unique embedding.

%todo: explaim
Besides the brand name, the commodity names are another key information for the products in the E-commerce platforms.
However, it is very hard to automatically check the the short title generate the commodity name correctly or not, since the source title usually contains multiple commodity name while they are not necessary to be the same as in background knowledge, like examples in Table~\ref{tab:cs}, 
So, we leave the test on commodity name in the manual evaluation in Section~\ref{sec:me}.

\subsection{Manual Evaluation}
\label{sec:me}
%todo: 增加对人的手工评价

\begin{table}
	\centering
	\caption{Manual evaluation results. Bold scores are the best. The improvements from baselines to MS-Pointer is statistically significant according to two-tailed t-test ($p < 0.05$).}
  \label{tab:manual}
	\setlength{\tabcolsep}{7pt}
	\begin{adjustbox}{max width=\linewidth}
	\begin{tabular}{l r c c c } \toprule
    Model & Accuracy & Comm. & Readability &  Info. \\ \midrule
   Seq2Seq-Gen & 83.67\% & 91.33\% & 4.54 &  3.73 \\
   Ptr-Gen & 91.03\% & 94.33\% & 4.71 & 4.09 \\ \midrule
   Trunc. & 29.67\% & 31.0\% & 2.67 & 2.31 \\
   TextRank & 5.67\% & 33.33\% & 2.56 & 2.77 \\
   LSTM-Del & 86.67\% & 92.33\% & 3.49 & 3.34 \\
   Seq2Seq-Del & 89.33\% & 93.67\%  & 3.52 & 3.46 \\
   Ptr-Net & 92.0\% &  94.67\%  & 4.79 & 4.21 \\
   Ptr-Concat & 94.33\% & 95.33\% & 4.81 & 4.31\\
   MS-Pointer & \textbf{97.33\%} & \textbf{98.0\%} & \textbf{4.87} & \textbf{4.55}\\ \midrule
   Human & 100\% & 100\% & 4.93 & 4.51 \\
    \bottomrule
  \end{tabular}
  \end{adjustbox}
\end{table}

Following the procedure in \cite{Filippova:Sentence:EMNLP2015,Tan:Abstractive:ACL2017}, we conduct manual evaluation on 300 random samples from our testset.
Three participants were asked to measure the quality of the short titles generated by each model from four perspectives:
\begin{enumerate*}[label=(\roman*)]
\item \textbf{Key Information Retention (Accuracy)}, is the key information properly kept in the short title?
\item \textbf{Commodity Name Retention (Comm.)}, is the commodity name correct?
\item \textbf{Readability}, how fluent, grammatical the short title
is?
\item \textbf{Informativeness (Info.)}, how informative the short title
is?
\end{enumerate*}

Comparing with conventional sentence summarization task, key information retention is an extra and essential requirement for product title summarization as we discussed in the introduction. %todo: polish
We use a very strict criteria for this property, the generated short title will be assessed as~1 only if it correctly retains both brand name and commodity name, otherwise~0.
In addition, we also test the commodity name retention precision solely as we explained in Section~\ref{sec:brand}.
The other two properties are assessed with a score from 1 (worst) to 5 (best).

The average results are presented in Table~\ref{tab:manual}.
Obviously, Trunc. and TextRank perform worst in this task as the same in automatic evaluations.
%much worse than other neural models since it cannot extract the informative words (\textit{e.g.}, commodity name like \textit{console} or \textit{running shoes}) from the tail of the source titles.
The results on \textit{Readability} metric show that all models built on Seq2Seq architecture (exclude Seq2Seq-Del) can generate very fluent titles.
They also verify our previous claim that it is difficult to produce a fluent title for the models based on the deletion.
The significant improvements on Accuracy metric demonstrate that our MS-Pointer can better retain the key information by the help of knowledge encoder.
Considering all three metrics, our MS-Pointer produces more readable and more informative titles, which shows the advantage of introducing the knowledge encoder.
Besides these baselines, we also conduct the experiments on human-written ground truths.
The results show that MS-Pointer performs very close with human, ever better on Info.~metric.
This is because the editors often produce very concise titles while MS-Pointer may produces longer titles with more information.

\subsection{Oneline A/B Testing}
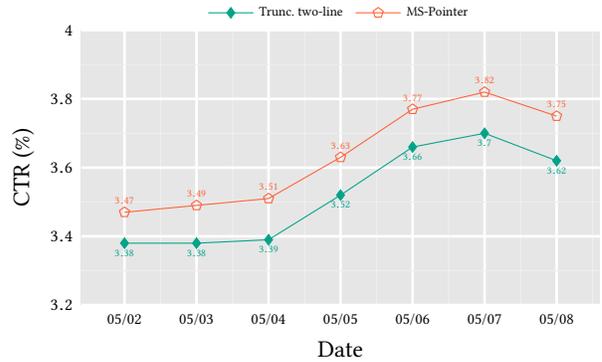
\begin{figure}
\pgfplotsset{
axis background/.style={fill=mercury},
grid=both,
  xtick pos=left,
  ytick pos=left,
  tick style={
    major grid style={style=white,line width=1pt},
    minor grid style=gallery,
    draw=none
    },
  minor tick num=1,
}
\centering
 \hspace{-5mm}\begin{tikzpicture}
    \begin{axis}[
      height=0.618\linewidth,
      width = \linewidth,
      xlabel=Date,
      ylabel=CTR (\%),
      xticklabels={05/02, 05/03, 05/04, 05/05, 05/06, 05/07, 05/08},
      xtick={1,2,3,4,5,6, 7},
      ymin=3.2, ymax=4.0,
      legend style={
          font=\tiny,
          draw=none,
          legend columns=-1,
          at={(0.5,1)},
          anchor=south,
          /tikz/every even column/.append style={column sep=0.9mm}
        },
        ymajorgrids,
        major grid style={draw=white},
        y axis line style={opacity=0},
        tickwidth=0pt,
        nodes near coords,
        every node near coord/.append style={anchor=north, font=\fontsize{4pt}{4pt}\selectfont},%\fontsize{4pt}{4pt}\selectfont
        legend entries = {Trunc. two-line, MS-Pointer},
        every tick label/.append style={font=\scriptsize}
        ]
      \addplot[color=mountain_meadow, mark=diamond*, every node near coord/.append style={yshift=0.3mm,anchor=north}] coordinates {
        (1, 3.38)
        (2, 3.38)
        (3, 3.39)
        (4, 3.52)
        (5, 3.66)
        (6, 3.70)
        (7, 3.62)
      };
      \addplot[color=tomato, mark=pentagon, every node near coord/.append style={anchor=south}] coordinates {
        (1, 3.47)
        (2, 3.49)
        (3, 3.51)
        (4, 3.63)
        (5, 3.77)
        (6, 3.82)
        (7, 3.75)
      };
    \end{axis}
  \end{tikzpicture}
  \caption{Online A/B Testing of CTR.}
  \label{fig:ab}
\end{figure}

Previous experimental results have shown the superiority of our proposed MS-Pointer.
%todo: taobao
In addition, we also deploy it in a real world application to test its practical performance.
This subsection presents the results of online evaluation in a recommendation scenario of Taobao mobile app with a standard A/B testing configuration.

%todo:实验设置
For online deployment, we generate the short titles for about 60 million products using our MS-Pointer model with 50 Tesla P100 GPUs in about 10 hours.
Due to the screen size limit, we restrict the length of the short title to 8--10 Chinese characters.
Note that the baseline deployed online is the truncated two-line titles (about 20 Chinese characters\footnote{The full titles usually contain about 30 Chinese characters}). %\footnote{This is also the }
The A/B testing system randomly split online users equally into two groups and direct them into two separate buckets (each bucket contains about 2 million daily active users) respectively.
Then for users in the bucket A (\textit{i.e.}, the baseline group), the titles they saw are the truncated two-line titles. While for users in the bucket B (\textit{i.e.}, the experimental group), the displayed short titles are generated by our MS-Pointer model.

\begin{table}
  \centering
  \caption{CTR improvements under different categories}
  \label{tab:ctr_cat}
 \begin{adjustbox}{max width=\linewidth}
  %\resizebox{0.5\textwidth}{!}{
  \begin{tabular}{c c c c } \toprule
   Clothing (Women) &  Shoes (Women) & Beauty & Electronics  \\ \midrule
   1.03\% & 2.71\% & 3.87\%  & 13.26\% \\ \bottomrule\toprule
     Clothing (Men) &  Shoes (Men) & Cell Phones & Computers  \\ \midrule
    5.84\% & 5.54\% & 9.75\% & 7.64\% \\
    \bottomrule
  \end{tabular}%}
  \end{adjustbox}
\end{table}

\begin{table*}
	\centering
	\caption{Examples of generated short titles. ``\textvisiblespace'' denotes visible whitespace in Chinese context.}
    \label{tab:cs}
	\begin{adjustbox}{max width=\textwidth}
	\renewcommand{\arraystretch}{1.2}
	\begin{tabular}{l l l } \toprule
     Original&  \begin{CJK*}{UTF8}{gkai}任天堂Switch\textvisiblespace 游戏机专用背夹电池MOD-X真皮保护套\end{CJK*} & \begin{CJK*}{UTF8}{gkai}美国\textvisiblespace 曼哈顿Manhattan\textvisiblespace Portage\textvisiblespace 邮差包\textvisiblespace 单肩包挎包 \end{CJK*}\\ %\textvisiblespace\textvisiblespace 正品现货
     Title & Nintendo Switch Console Dedicated Battery Case MOD-X Leather Case & US Manhattan Manhattan Portage Messenger Bag Shoulder Bag Satchel \\ \midrule %genuine in stock
     Background & \textsc{Brand Name}: MOD-X & \textsc{Brand Name}: \begin{CJK*}{UTF8}{gkai}Manhattan Portage \end{CJK*} \\ 
     Knowledge & \textsc{Commodity Name}: \begin{CJK*}{UTF8}{gkai}电池\end{CJK*} // Battery & \textsc{Commodity Name}: \begin{CJK*}{UTF8}{gkai}背提包\end{CJK*} // Handbag and Knapsack \\ \midrule %
     Ground Truth & \begin{CJK*}{UTF8}{gkai}MOD-X任天堂Switch背夹电池\end{CJK*} // MOD-X Nintendo Switch Battery Case  &  \begin{CJK*}{UTF8}{gkai}Manhattan\textvisiblespace Portage\textvisiblespace 邮差包 \end{CJK*} // Manhattan Portage Messenger Bag\\ \midrule
    Trunc. & \begin{CJK*}{UTF8}{gkai}任天堂Switch\textvisiblespace 游戏机专用\end{CJK*} // Nintendo Switch Console Dedicated & \begin{CJK*}{UTF8}{gkai}美国\textvisiblespace 曼哈顿Manhattan\textvisiblespace Portage\textvisiblespace\end{CJK*} // US Manhattan Manhattan Portage \\
   TextRank & \begin{CJK*}{UTF8}{gkai}游戏机专用背夹电池MOD-X\end{CJK*} // Console Dedicated Battery Case MOD-X & \begin{CJK*}{UTF8}{gkai}Portage邮差包单\end{CJK*} // Portage Messenger Bag single\\
   Seq2Seq-Gen & \begin{CJK*}{UTF8}{gkai}任天堂NS\textvisiblespace switch主机\end{CJK*} // Nintendo NS Switch Console & \begin{CJK*}{UTF8}{gkai}ORSLOW单肩斜挎包\end{CJK*} // ORSLOW Shoulder Crossbody Bag \\
   Seq2Seq-Del & \begin{CJK*}{UTF8}{gkai}任天堂游戏机保护套\end{CJK*} // Nintendo Console Case &  \begin{CJK*}{UTF8}{gkai}曼哈顿Manhattan\textvisiblespace Portage\textvisiblespace 差包包\end{CJK*} // Manhattan Manhattan Portage Bag Bag\\
   LSTM-Del & \begin{CJK*}{UTF8}{gkai}任天堂游戏机\end{CJK*} // Nintendo Console & \begin{CJK*}{UTF8}{gkai}曼哈顿Manhattan\textvisiblespace Portage包\textvisiblespace 包包\end{CJK*} // Manhattan Manhattan Portage Bag Bag\\
   Ptr-Gen & \begin{CJK*}{UTF8}{gkai}任天堂switch游戏机\end{CJK*} // Nintendo Switch Console& \begin{CJK*}{UTF8}{gkai}美国曼哈顿Manhattan\textvisiblespace 邮差包\end{CJK*} // US Manhattan Manhattan Messenger Bag\\ \midrule
   Ptr-Net & \begin{CJK*}{UTF8}{gkai}任天堂switch游戏机\end{CJK*} // Nintendo Switch Console  & \begin{CJK*}{UTF8}{gkai}曼哈顿单肩包\end{CJK*} // Manhattan Shoulder Bag\\
   Ptr-Concat   & \begin{CJK*}{UTF8}{gkai}MOD-X任天堂Switch背夹电池\end{CJK*}  // MOD-X Nintendo Switch Battery Case & \begin{CJK*}{UTF8}{gkai}美国Manhattan\textvisiblespace Portage单肩包\end{CJK*} // US Manhattan Portage Shoulder Bag\\
   MS-Pointer & \begin{CJK*}{UTF8}{gkai}MOD-X任天堂Switch背夹电池\end{CJK*} // MOD-X Nintendo Switch Battery Case & \begin{CJK*}{UTF8}{gkai}美国Manhattan\textvisiblespace Portage\textvisiblespace 邮差包\end{CJK*} // US Manhattan Portage Messenger Bag\\
    \bottomrule
  \end{tabular}
  \end{adjustbox}
\end{table*}

%todo:polish
This online A/B testing lasted for one week.
All the settings of the two buckets are identical except the displayed titles.
We adopt the Click-Through Rate (CTR) to measure the performance since the product titles often are a crucial decision factor in determining
whether to click a product or skip to another.
It is calculated as:
\begin{equation*}
	\mathrm{CTR} = \frac{\# \mathrm{click}}{\# \mathrm{impression}}
\end{equation*}
where \#$\mathrm{click}$ is the number of clicks on the product, \#$\mathrm{impression}$ is the number of times the product is shown.

Figure~\ref{fig:ab} shows the results of overall CTR for all products in the two buckets in one week (from 05/02/2018 to 05/08/2018).
It is obvious that the experimental bucket (\textit{i.e.}, MS-Pointer) significantly outperforms the baseline bucket ($p < 0.05$).
This clearly shows that the single-line short titles generated by MS-Pointer are more user-friendly and more likely to attract users to click on products.
%todo:丰富

In order to gain some intuition on how the these generated short titles affect the users, we also analysis the CTR improvements under different categories.
Table~\ref{tab:ctr_cat} shows the overall CTR improvements under some typical categories in one week.
This table reveals an interesting result that CTR improvements on categories like electronic devices are significantly higher than categories like clothing and beauty, especially on clothing(women).
This may be caused by the users' different behavioral pattern under different categories.
When browsing the products like women's clothing or beauty, users usually pay more attention to the modifier words (\textit{e.g.}, \begin{CJK*}{UTF8}{gkai}刺绣\end{CJK*}/\textit{embroidery}, \begin{CJK*}{UTF8}{gkai}丝绸\end{CJK*}/\textit{silk}, \begin{CJK*}{UTF8}{gkai}高腰\end{CJK*}/\textit{high-waist}, or \begin{CJK*}{UTF8}{gkai}水润\end{CJK*}/\textit{moisturizing}).
However, ten Chinese characters are often difficult to include all the modifiers that attract the customers under these categories.
While browsing the electronic devices or men's clothes, users usually do not care about these modifier words.
A short and clear title is more attractive to users under these categories.

\subsection{Case Study}

%美国 曼哈顿Manhattan Portage 邮差包 单肩包挎包  正品现货
%GOYEAH 苹果笔记本电脑包13寸15寸公文包手提单肩男士出差旅行包
%日本专柜mosh合作Kitty 爱丽丝限定不锈钢保温保冷杯 牛奶杯
% Catalyst 苹果手表apple watch3防水保护套iwatch2硅胶表带壳运动
%任天堂 Nintendo Switch NS PRO手柄 专用经典手柄 原装正品 现货
% ANKER 移动电源20000毫安大容量充电宝适用新Macbook任天堂Switch
% 任天堂Switch 游戏机专用背夹电池移动电源MOD-X真皮保护套

%we show some typical examples of all systems.
To better understand what can be learned by each model, we show some typical examples they generated in Table~\ref{tab:cs}.
As expected, Trunc.~and TextRank perform worst on these two cases.
In terms of readability, all models based on encoder-decoder architecture can produce fluent and grammatical short titles.
However, in terms of informativeness, MS-Pointer and Ptr-Concat perform much better than other baseline, thanks to introducing the background knowledge.
Moreover, the short titles generated by our MS-Pointer are comparable with those ground truths.

With regard to informativeness, the difficult cases are those where brand name does not appear in the head of the title and where the baselines still try to pick it out from the head as they learned in the training data. %90%以上在前面
For the right case in Table~\ref{tab:cs}, Seq2Seq-Gen generates a wrong brand name ``\textit{ORSLOW}''; Ptr-Gen and Ptr-Net also fail to keep the brand intact.
This type of error is unacceptable in the real-word applications.
Nevertheless, our MS-Pointer surpasses Ptr-Concat with a more informative commodity name ``\textit{Messenger Bag}''.
The left case in Table~\ref{tab:cs} is more tough since it contains several brand name (\begin{CJK*}{UTF8}{gkai}任天堂\end{CJK*}/\textit{Nintendo} and \textit{MOD-X}) and commodity name (\begin{CJK*}{UTF8}{gkai}游戏机\end{CJK*}/\textit{Console},\begin{CJK*}{UTF8}{gkai}背夹电池\end{CJK*}/\textit{Battery Case}, and \begin{CJK*}{UTF8}{gkai}保护套\end{CJK*}/\textit{Case}).
Ptr-Net, Ptr-Gen, and deletion based methods all fail to generate the brand name ``\textit{MOD-X}'' and commodity name ``\begin{CJK*}{UTF8}{gkai}背夹电池\end{CJK*}/\textit{Battery Case}''.
It is not easy to copy the right brand name and commodity name for models without the background knowledge.

\tikzset{
  tit/.style = {text depth=.25ex, text height=1.5ex, inner sep=0, outer sep=0}
}
\pgfplotsset{
%axis background/.style={fill=none},
colormap={my_colormap}{
    rgb255=(56, 113, 66),
    rgb255=(239, 251, 242),
},
}
\pgfplotsset{
    colormap default colorspace=rgb,
    colormap={blueColormap}{
       color(0000pt)=(cosmic_latte!50!white);
       color(1000pt)=(chinook!50!cosmic_latte);
       color(1500pt)=(chinook);
       color(2500pt)=(pastel_green);
       %color(4000pt)=(pastel_green);
       color(4000pt)=(ocean_green);
       color(5000pt)=(killarney);
   }
}

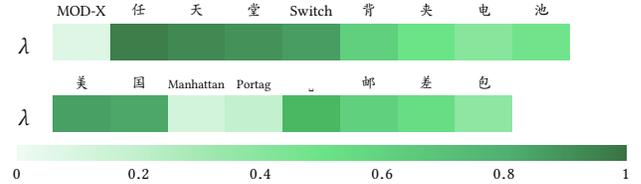
\begin{figure}
\centering
\resizebox{\linewidth}{!}{
\begin{tikzpicture}[]
\tikzstyle{every node}=[font=\scriptsize]
\fill[cosmic_latte] (0, 0) rectangle ++(0.8,0.5);
\fill[killarney!80!ocean_green] (0.8, 0) rectangle ++(0.8,0.5);
\fill[killarney!62!ocean_green] (1.6, 0) rectangle ++(0.8,0.5);
\fill[killarney!47!ocean_green] (2.4, 0) rectangle ++(0.8,0.5);
\fill[killarney!30!ocean_green] (3.2, 0) rectangle ++(0.8,0.5);
\fill[ocean_green!40!pastel_green] (4, 0) rectangle ++(0.8,0.5);
\fill[pastel_green] (4.8, 0) rectangle ++(0.8,0.5);
\fill[pastel_green!50!chinook] (5.6, 0) rectangle ++(0.8,0.5);
\fill[pastel_green!90!chinook] (6.4, 0) rectangle ++(0.8,0.5);

\node[] (w1) at (-0.4, 0.2) {\normalsize$\lambda$};
\node[tit] (w1) at (0.4, 0.7) {MOD-X};
\node[tit] (w2) at (1.2, 0.7) {\begin{CJK*}{UTF8}{gkai}任\end{CJK*}};
\node[tit] (w3) at (2, 0.7) {\begin{CJK*}{UTF8}{gkai}天\end{CJK*}};
\node[tit] (w4) at (2.8, 0.7) {\begin{CJK*}{UTF8}{gkai}堂\end{CJK*}};
\node[tit] (w5) at (3.6, 0.7) {Switch};
\node[tit] (w6) at (4.4, 0.7) {\begin{CJK*}{UTF8}{gkai}背\end{CJK*}};
\node[tit] (w7) at (5.2, 0.7) {\begin{CJK*}{UTF8}{gkai}夹\end{CJK*}};
\node[tit] (w8) at (6, 0.7) {\begin{CJK*}{UTF8}{gkai}电\end{CJK*}};
\node[tit] (w9) at (6.8, 0.7) {\begin{CJK*}{UTF8}{gkai}池\end{CJK*}};

\node[] (w1) at (-0.4, -0.8) {\normalsize$\lambda$};
\fill[killarney!25!ocean_green] (0, -1) rectangle ++(0.8,0.5);
\fill[killarney!15!ocean_green] (0.8, -1) rectangle ++(0.8,0.5);
\fill[cosmic_latte!80!chinook] (1.6, -1) rectangle ++(0.8,0.5);
\fill[cosmic_latte!55!chinook] (2.4, -1) rectangle ++(0.8,0.5);
\fill[chateau_green!50!ocean_green] (3.2, -1) rectangle ++(0.8,0.5);
\fill[ocean_green!37!pastel_green] (4, -1) rectangle ++(0.8,0.5);
\fill[ocean_green!12!pastel_green] (4.8, -1) rectangle ++(0.8,0.5);
\fill[pastel_green!38!chinook] (5.6, -1) rectangle ++(0.8,0.5);

\node[tit] (t1) at (0.4, -0.3) {\begin{CJK*}{UTF8}{gkai}美\end{CJK*}};
\node[tit] (t2) at (1.2, -0.3) {\begin{CJK*}{UTF8}{gkai}国\end{CJK*}};
\node[tit] (t3) at (2, -0.3) {\tiny Manhattan};
\node[tit] (t4) at (2.8, -0.3) {\tiny Portag};
\node[tit] (t5) at (3.6, -0.3) {\textvisiblespace};
\node[tit] (t6) at (4.4, -0.3) {\begin{CJK*}{UTF8}{gkai}邮\end{CJK*}};
\node[tit] (t7) at (5.2, -0.3) {\begin{CJK*}{UTF8}{gkai}差\end{CJK*}};
\node[tit] (t8) at (6, -0.3) {\begin{CJK*}{UTF8}{gkai}包\end{CJK*}};

\begin{axis}[
    hide axis,
    scale only axis,
    yshift=-1.5cm,
    xshift=-0.5cm,
    colorbar,
    colorbar horizontal,
    %colorbar sampled,
    colorbar style={
        %rotate=90,
        %anchor=south east,
        height=0.2cm,
        width=\linewidth,
        axis line style={draw=none},
    },
    tick style = {draw=none},
    point meta min=0, 
    point meta max=1
    ]
%\addplot [draw=none] coordinates {(0,0)};

\end{axis}
        
\end{tikzpicture}}
\caption{Heatmap of $\lambda$ for the cases in Table~\ref{tab:cs}.}
\label{fig:cs}
\end{figure}
We also visualize the gating weight $\lambda$ in Figure~\ref{fig:cs} for each step of the decoder in these two cases.
It is easy to see that our MS-Pointer produces very small $\lambda$ for the brand name (\textit{e.g.}, \textit{MOD-X} and \textit{Manhattan Portag}) to force the model to copy them from the knowledge encoder, and larger $\lambda$ to force the model to copy the modifier from the source title encoder.

%比较难的case，是品牌不出现头部，但是model在训练数据中观察到的主要是头部

%todo: 可视化图\lambda？

\section{Conclusion} % and Future Work

In this paper, we study the product title summarization problem in E-commerce.
In response to two particular constraints in this task, we propose a novel model MS-Pointer to generate the short titles by copying words from not only the source title but also the background knowledge.
We perform extensive experiments with realistic data from a popular Chinese E-commerce website.
%Empirical study on both automatic evaluation metrics and human annotations shows
Experimental results demonstrate that our model can generate informative and fluent short titles and significantly outperform other strong baselines.
Finally, online A/B testing shows the significant business impact of our model in a real-world application.

%In fact, our proposed model is very close to humans on automatic metrics.
%Finally, online A/B testing shows that the short titles produced by our MS-Pointer model can siginifi improve the CTR of the products.

%deployment of our MS-Pointer model on a popular E-commerce mobile app has yielded a significant business impact, as measured by the click-through rate.

Although introducing background knowledge makes our proposed model significantly outperforms the other baselines, the knowledge used in this paper is very elementary and limited.
An interesting direction for the future work would be how to incorporate the  knowledge graph into pointer network for product title summarization.
Another direction is how to produce personalized short titles for different users, considering they may care about different properties about the products.

\section*{Acknowledgments}

We thank Siying Li, Yixing Fan, Liang Pang, Wen Chen, and Yanran Li for valuable comments and discussion.
We also thank Abigail See for his kindness in sharing source code.


%\bibliographystyle{ACM-Reference-Format}
%\bibliography{cikm2018}

\end{document}